\documentclass[preprint,5p,times,authoryear]{elsarticle}
\usepackage{amssymb}
\usepackage{amsmath}
\usepackage[switch]{lineno}
\usepackage{tikz}
\usepackage{pgfplots}
\pgfplotsset{compat=1.18}
\usepackage[caption=false,font=footnotesize,labelfont=rm,textfont=rm]{subfig}
\usepackage{caption}
\usepackage{multirow}
\usepackage{booktabs}
\usepackage{makecell}
\usepackage{booktabs}
\DeclareCaptionFormat{cont}{#1 (cont.)#2#3\par}
\journal{ISPRS Journal of Photogrammetry and Remote Sensing}
\usepackage{url}
\usepackage{hyperref}
\hypersetup{
    colorlinks=true,
    linkcolor=blue,
    filecolor=magenta,      
    urlcolor=cyan,
    pdftitle={Think2Seg-RS},
    pdfpagemode=FullScreen,
    }
\usepackage[capitalise]{cleveref}
\makeatletter
\renewcommand{\fnum@figure}{Fig. \thefigure}
\makeatother

\usepackage{microtype}
\usepackage{listings}
\usepackage{changes}
\usepackage{xcolor}
\usepackage[normalem]{ulem}
\usepackage{threeparttable}
\usepackage{colortbl}

\definecolor{addcolor}{RGB}{255,0,255}    % Magenta
\definecolor{delcolor}{RGB}{255,165,0}   % Orange
\definecolor{replcolor}{RGB}{138,43,226} % Vivid Violet

\makeatletter
\AddToHook{cmd/added/before}{\def\Changes@AuthorColor{addcolor}}
\AddToHook{cmd/deleted/before}{\def\Changes@AuthorColor{delcolor}}
\AddToHook{cmd/replaced/before}{\def\Changes@AuthorColor{replcolor}}
\makeatother

\begin{document}
% \setpagewiselinenumbers
% \linenumbers

\begin{frontmatter}



\title{Bridging Semantics and Geometry: A Decoupled LVLM–SAM Framework for Reasoning Segmentation in Optical Remote Sensing
}

\author{Xu Zhang}
\author{Junyao Ge}
\author{Yang Zheng}
\author{Kaitai Guo}
\author{Jimin Liang\corref{cor1}} %% Author name
\ead{jimleung@mail.xidian.edu.cn}
% \ead{junyao_ge@stu.xidian.edu.cn, zhengy@xidian.edu.cn, ktguo@xidian.edu.cn,  jimleung@mail.xidian.edu.cn}
\cortext[cor1]{Corresponding author.}

% \affiliation{organization={School of Electronic Engineering, Xidian University},
%             city={Xi'an},
%             state={Shaanxi 710071},
%             country={China}}

\address{School of Electronic Engineering, Xidian University, Xi'an, Shaanxi 710071, China}

\begin{abstract}
Large Vision–Language Models (LVLMs) hold great promise for advancing optical remote sensing (RS) analysis, yet existing reasoning segmentation frameworks couple linguistic reasoning and pixel prediction through end-to-end supervised fine-tuning, leading to weak geometric grounding and limited generalization across tasks. 
To address this, we developed Think2Seg-RS, a decoupled framework that trains an LVLM prompter to control a frozen Segment Anything Model (SAM) via structured geometric prompts. 
Through a mask-only Group Relative Policy Optimization (GRPO) reinforcement learning objective driven strictly by final mask IoU, the LVLM learns to translate abstract semantic reasoning into spatially grounded actions, achieving state-of-the-art performance on the EarthReason dataset.
Notably, Think2Seg-RS outperforms leading approaches such as RemoteReasoner and SegEarth-R1 on the EarthReason dataset by reaching a test cIoU of 75.60\% and gIoU of 73.36\%, yielding absolute improvements of 6.47\% and 2.40\% over the strongest baseline, respectively.
Zero-shot evaluations across three referring segmentation benchmarks reveal a fundamental distinction in task inductive bias, exposing a distinct divide between semantic-level grounding---which aggregates all regions matching a conceptual intent---and instance-level tasks that demand discrete object separation. 
We further found that compact segmenters outperform larger ones under semantic-level supervision by mitigating textural over-segmentation, and that unconstrained negative prompting is unstable in heterogeneous aerial backgrounds.
Together, these findings demonstrate that optimizing LVLMs through direct segmentation feedback offers a scalable framework for complex geospatial reasoning, effectively bridging the gap between abstract language understanding and precise pixel-level execution.
Our code and model are available at \url{https://github.com/Ricardo-XZ/Think2Seg-RS}.
\end{abstract}


% \begin{highlights}
% \end{highlights}

%% Keywords

\end{frontmatter}




\section{Introduction}
\label{sec:intro}

\begin{figure*}[ht]
    \centering
    \includegraphics[width=0.95\textwidth]{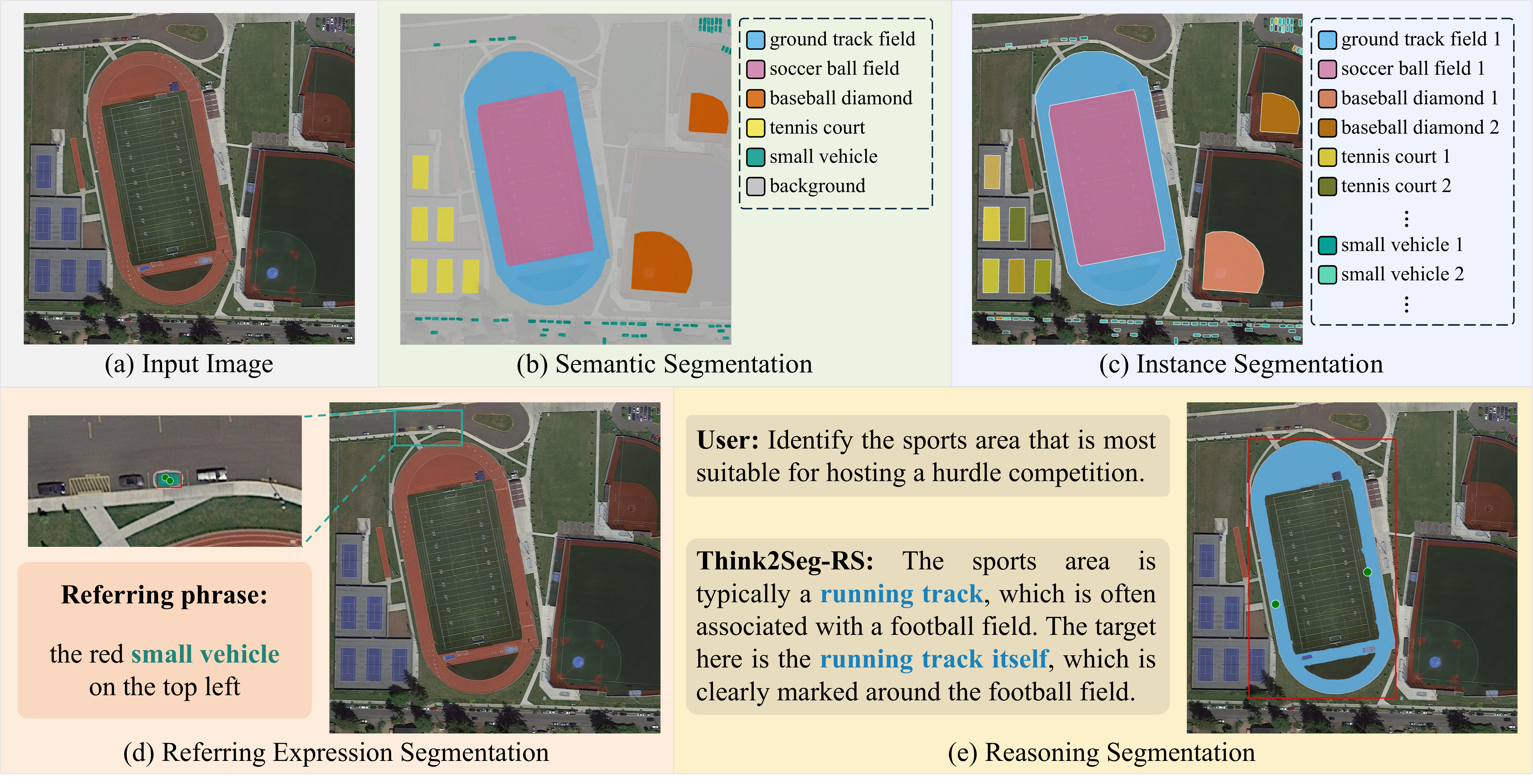}
    \caption{Illustration of the evolution of segmentation paradigms in RS imagery. 
    (a) Input image sampled from the iSAID dataset~\citep{waqas2019isaid}. 
    (b) Semantic segmentation ground truth provided by the dataset, assigning fixed class labels to every pixel. 
    (c) Instance segmentation ground truth, distinguishing individual objects of the same class. 
    (d) Referring expression segmentation result (mask and generated prompts,box in red and positive points in green) produced by our proposed Think2Seg-RS method, guided by explicit language queries. 
    (e) Reasoning segmentation result by Think2Seg-RS, handling implicit, compositional queries.}
    \label{fig:Segmentation-evolution}
\end{figure*}

Segmentation of remote sensing (RS) imagery has long been recognized as a cornerstone of intelligent image interpretation. Over the years, segmentation techniques have evolved from \emph{semantic segmentation}, which assigns a fixed class label to every pixel \citep{yuan2021review,lv2023deep,guo2019learnable,ma2024sam-assisted}, to \emph{instance segmentation}, which further distinguishes individual objects of the same category \citep{chen2024rsprompter,xu2021swin-instance,liu2024learning,su2020hq}, with illustrative examples provided in Fig.~\ref{fig:Segmentation-evolution}. While these approaches have achieved notable progress, existing paradigms remain inherently constrained by predefined label sets and therefore struggle to cope with open-world conditions where user needs cannot be fully anticipated. 

% To overcome this rigidity, \emph{referring expression segmentation} is proposed \citep{rrsis,liu2024rrsis-d,lei2025exploring,pan2024rethinking,yao2025remotesam,dong2024cross}.
% \added{To overcome this rigidity, \emph{referring expression segmentation} was introduced \citep{rrsis}, alongside specialized benchmark datasets \citep{liu2024rrsis-d,dong2024cross,yao2025remotesam} and cross-modal architectures \citep{pan2024rethinking,lei2025exploring}.}
To overcome this rigidity, \emph{referring expression segmentation} was introduced \citep{rrsis}, with efforts spanning specialized benchmarks \citep{liu2024rrsis-d,dong2024cross,yao2025remotesam} and cross-modal alignment architectures \citep{lei2025exploring,pan2024rethinking}.
Formally, this task focuses on explicit visual delineation, where the user query directly specifies the target through observable attributes such as category names, colors, or spatial relationships (e.g., ``the red small vehicle on the top left,'' see Fig.~\ref{fig:Segmentation-evolution}).
However, real-world queries are often implicit and compositional, necessitating the emerging paradigm of \emph{reasoning segmentation}~\citep{li2025segearth,yao2025remotereasoner}. 
This task is defined as the delineation of targets specified by implicit queries, where identification requires complex reasoning involving spatial context, causal relations, or domain knowledge.
For instance, addressing queries like ``Identify the sports area that is most suitable for hosting a hurdle competition'' requires models that can go beyond visual attribute matching to simultaneously (i) interpret implicit natural language instructions, (ii) reason about spatial logic and domain semantics, and (iii) deliver precise pixel-level delineations.
This progression from explicit referring to implicit reasoning represents the latest stage in RS segmentation methods.

Although Large Vision–Language Models (LVLMs) have demonstrated impressive instruction following and multimodal reasoning on natural images, directly applying them to RS is far from sufficient. In terms of training data and objectives, mainstream LVLMs (e.g., LLaVA-1.5~\citep{liu2023improvedllava}, InternVL3~\citep{zhu2025internvl3}, and Qwen-2.5-VL~\citep{bai2025qwen2}, which are the open-source models we adopt in our study) are primarily optimized for image--text instruction following, visual question answering, document understanding, and grounding with boxes, rather than dense, pixel-accurate segmentation masks. This mismatch yields domain gaps (e.g., nadir views, tiny objects, repetitive textures, and geographic semantics that differ markedly from the natural images dominating current LVLM training corpora) and task gaps (e.g., from captioning/question-answering or box-level grounding to fine-grained mask delineation). Even if an LVLM correctly parses an implicit query, converting its reasoning into reliable, fine-grained masks remains nontrivial because current LVLM heads and training signals are not tailored for precise spatial delineation. 
Moreover, early RS reasoning-segmentation pipelines that map compressed LVLM embeddings directly to masks with a monolithic decoder~\citep{li2025segearth} suffer from an intrinsic information bottleneck. Specifically, attempting to encode high-frequency visual details into compact semantic embeddings (e.g., a single \texttt{<SEG>} token) inevitably leads to the loss of fine-grained spatial fidelity required for precise RS object delineation, while the rigid monolithic design restricts architectural scalability and the opaque implicit decoding process limits interpretability.

To address these challenges, we adopt a decoupled reasoning–execution paradigm that explicitly separates reasoning from segmentation. In this design, an LVLM serves as the reasoning module, interpreting complex or implicit instructions and emitting structured geometric prompts (bounding boxes and points), while a universal segmenter acts as the execution module to translate prompts into masks. 
Crucially, by transmitting reasoning outcomes via explicit geometric coordinates rather than compressed latent vectors, this design bypasses the information bottleneck, allowing the segmenter to leverage its high-fidelity internal features for precise pixel-level delineation.
The Segment Anything Models (SAMs) \citep{kirillov2023segment,ravisam} are particularly suited for this role due to their strong zero-shot generalization and promptable interfaces; their RS variants already show robust performance \citep{ren2024segmentanythingfromspace,wang2023samrs,zheng2024segment}.

The practical bottleneck in RS therefore lies not in the segmentation capacity of SAM itself, but in how to generate suitable prompts that bridge high-level reasoning with low-level execution. Concretely, this involves (i) \emph{semantic grounding}—mapping implicit, compositional queries to the correct subset of instances and their spatial extents; (ii) \emph{spatial precision and instance control}—deciding where and how prompts (e.g., boxes and points) should be issued so that SAM yields accurate masks under scale variation, clutter, and ambiguous boundaries; and (iii) \emph{automation at scale}—eliminating the brittleness and cost of manual prompt design. Our solution is to train the LVLM to automatically generate such high-quality prompts that faithfully reflect both user intent and RS scene context, while keeping SAM frozen to preserve its universality and avoid domain-specific retraining.

Building on this idea, we present Think2Seg-RS, a training-efficient framework for RS reasoning segmentation. In our design, the LVLM (Qwen-2.5-VL family) is the only trainable component and is optimized to produce well-formed JSON-structured prompts (a configuration validated by our ablation study: a bounding box for localization and two positive points for refinement) from image–text inputs, while a frozen SAM2 \citep{ravisam} executes segmentation conditioned on these prompts. To bridge the reasoning--segmentation gap, we adopt Group Relative Policy Optimization (GRPO) \citep{shao2024deepseekmath} with a simple, result-oriented reward composed of (i) an output-format reward (ensuring valid \verb|<think>|/\verb|<answer>| tags and JSON) and (ii) the final Intersection-over-Union (IoU) of the SAM-produced mask. This establishes a direct feedback loop from segmentation outcomes to prompt generation, enabling the LVLM to learn a reasoning-to-segmentation skill that generalizes beyond handcrafted prompting strategies and yields a mask-only, modular training recipe.

Recent concurrent works such as Seg-Zero~\citep{liu2025seg}, VisionReasoner~\citep{liu2025visionreasoner}, and RemoteReasoner~\citep{yao2025remotereasoner} also adopt decoupled LVLM--SAM paradigms with reinforcement learning to generate prompts for segmentation. However, these approaches depend on pseudo box--point supervision, which constrains flexibility and adds annotation overhead. 
More critically, such proxy supervision forces the model to mimic heuristic geometric rules (e.g., bounding box tightness or centroid placement) that do not necessarily guarantee the best segmentation outcomes from the frozen SAM. By contrast, Think2Seg-RS employs a mask-only optimization scheme to bypass these intermediate proxies and directly optimize the final segmentation quality through execution feedback from SAM. This paradigm enlarges the search space for prompting strategies, enabling the LVLM to discover functional prompting patterns that are better aligned with SAM's behavior than handcrafted heuristics, while simultaneously reducing annotation cost and offering a more scalable alternative.

In summary, our contributions are threefold: 
\begin{itemize}
    \item \textbf{Think2Seg-RS framework.} We present Think2Seg-RS, which adopts a decoupled reasoning--execution design to separate high-level language understanding and reasoning (LVLM) from low-level pixel execution (SAM). This design eliminates manual prompt engineering and domain-specific fine-tuning, while preserving modularity and plug-and-play adaptability to future LVLMs.
    
    \item \textbf{Mask-only GRPO optimization.} We develop a lightweight optimization scheme based on GRPO that uses only mask annotations and a simple IoU + format reward, avoiding costly instance-level box--point supervision and equipping the LVLM with a robust reasoning-to-prompt capability.
    
    \item \textbf{SOTA accuracy and transfer.} Think2Seg-RS sets new state-of-the-art (SOTA) results on the EarthReason reasoning segmentation dataset~\citep{li2025segearth} and exhibits strong zero-shot transfer to the RRSIS-D~\citep{liu2024rrsis-d}, RISBench~\citep{dong2024cross} and RefSegRS~\citep{rrsis} referring expression segmentation benchmarks, evidencing that the learned prompting policy generalizes across datasets and task formulations rather than overfitting to dataset-specific query patterns.
\end{itemize}

\section{Related Work}

\subsection{From Semantic Labels to Reasoning Segmentation in RS}

The evolution of RS image segmentation has followed a clear trajectory. Early work was dominated by semantic \citep{yuan2021review,lv2023deep} and instance segmentation \citep{xu2021swin-instance,su2020hq}, supported by the release of large-scale RS benchmarks \citep{waqas2019isaid,loveda,azimi2019skyscapes,wang2023samrs}.
For example, specific approaches within these paradigms refine target delineation by aggregating multi-scale context \citep{liu2024learning}, enforcing boundary constraints \citep{ma2024sam-assisted}, or leveraging prompt-driven mechanisms \citep{chen2024rsprompter}.
These methods established strong baselines but remain inherently tied to predefined label sets, limiting their adaptability to open-world scenarios. This challenge has spurred a shift toward more flexible, language-guided paradigms for interactive querying.

% original
% The evolution of RS image segmentation has followed a clear trajectory. Early work was dominated by semantic \citep{lv2023deep,yuan2021review,guo2019learnable,ma2024sam-assisted} and instance segmentation \citep{chen2024rsprompter,xu2021swin-instance,liu2024learning,su2020hq}, supported by the release of large-scale RS benchmarks \citep{waqas2019isaid,loveda,wang2023samrs,azimi2019skyscapes}. These methods established strong baselines but remain inherently tied to predefined label sets, limiting their adaptability to open-world scenarios. This challenge has spurred a shift toward more flexible, language-guided paradigms for interactive querying.

Referring expression segmentation represents one such direction, enabling users to locate targets through simple free-form text descriptions. \cite{rrsis} first introduced this paradigm into remote sensing, formally defining the task as referring remote sensing image segmentation (RRSIS) and constructing the RefSegRS benchmark dataset to support it. Building on this foundation, \cite{liu2024rrsis-d} proposed the RRSIS-D dataset and a Rotated Multi-Scale Interaction Network (RMSIN), which integrates multi-scale interaction and adaptive rotated convolution to better capture the large spatial variations and diverse orientations of objects in aerial scenes. More recently, \cite{pan2024rethinking} identified limitations of the dominant implicit optimization paradigm in RRSIS, namely inter-domain misalignment and semantics-agnostic prediction, and addressed them with a Dual Alignment Network (DANet) that combines explicit affinity alignment and a reliable agent alignment module to enhance cross-modal consistency. 

Despite these advances, existing referring expression segmentation methods are fundamentally tailored to explicit expressions, and their alignment strategies remain insufficient for handling implicit, compositional queries that demand deeper reasoning about spatial relationships, causal logic, and geographic commonsense. This gap has motivated the emerging direction of reasoning segmentation. Early explorations have appeared in both the natural image and remote sensing communities. \cite{lai2024lisa} introduced the concept of \emph{reasoning segmentation} for natural images, proposing the large Language Instructed Segmentation Assistant (LISA) framework to infer segmentation masks from implicit, compositional language queries. Building on this idea, \cite{li2025segearth} extended the paradigm to the geospatial domain with SegEarth-R1, which pioneered the task of \emph{geospatial pixel reasoning} and released the EarthReason benchmark as the first dedicated dataset for this problem. In this paper, we use the terms \emph{reasoning segmentation} and \emph{geospatial pixel reasoning} interchangeably, as they denote the same underlying challenge of linking high-level reasoning with fine-grained pixel delineation in RS imagery.

Crucially, beyond the complexity of queries, these paradigms differ fundamentally in annotation granularity. Existing benchmarks bifurcate into two categories: instance-level datasets like RRSIS-D~\citep{liu2024rrsis-d} and RISBench~\citep{dong2024cross}, which emphasize the separation of discrete objects; and semantic-level datasets like RefSegRS~\citep{rrsis} and EarthReason~\citep{li2025segearth}, where the ground truth aggregates all regions matching a concept. To accommodate these diverse granularities, prior works have focused on explicit unification. For example, MaskFormer~\citep{cheng2021per} unifies semantic and instance tasks via a mask classification architecture, while multimodal frameworks like EVF-SAM~\citep{zhang2024evf} and GSVA~\citep{xia2024GSVA} resolve granularity conflicts through joint training, employing specific identifiers (e.g., a \texttt{[semantic]} prefix) or learning to enumerate multiple \texttt{[SEG]} tokens to cover diverse targets. However, these strategies necessitate simultaneous supervision from both instance-level and semantic-level annotations. Consequently, the challenge of granularity mismatch remains underexplored, particularly in zero-shot scenarios where a model trained on one granularity must implicitly adapt to the other. This misalignment often induces conflicting inductive biases when the evaluation protocol differs from the training supervision.

\subsection{Universal Execution Engines: SAM for Remote Sensing}
The transition from explicit grounding to implicit reasoning, further complicated by diverse annotation granularities, necessitates a flexible model capable of generalizing beyond fixed taxonomies.
The Segment Anything Model (SAM) \citep{kirillov2023segment} aligns well with this need: as a promptable, category-agnostic segmenter, it accepts simple geometric cues (points, boxes, or masks) and exhibits strong zero-shot generalization, thereby reshaping interactive segmentation workflows.

These capabilities have been rapidly explored in the RS domain. \cite{wang2023samrs} introduced SAMRS, a large-scale RS segmentation corpus constructed by leveraging SAM on existing detection datasets, showing how SAM can bootstrap pixel-level labels at scale. Independent evaluations reported that SAM maintains competitive zero-shot behavior across multiple overhead-imagery benchmarks, further indicating its cross-domain robustness in RS scenes \citep{ren2024segmentanythingfromspace}. Beyond static single-image settings, \cite{zheng2024segment} adapted SAM for bitemporal inputs and formulated Segment Any Change, demonstrating training-free zero-shot change detection via latent-space matching and point queries. Collectively, these studies highlighted SAM’s effectiveness and portability for RS tasks, from single-image segmentation to change detection.

However, SAM’s performance hinges on the quality of its input prompts, creating a bottleneck for large-scale or automated RS pipelines. Recent work therefore focuses on automated prompting. \cite{chen2024rsprompter} proposed RSPrompter, which generates prompt embeddings so that SAM can produce semantically discriminative instance masks in RS imagery. In parallel, \cite{zhang2023text2seg} presented Text2Seg, coupling Grounding~DINO for text-to-box grounding with CLIP-derived point cues to steer SAM via language queries. \cite{osco2023segment} systematically evaluated SAM for RS applications and discussed practical recipes (e.g., one-shot/lightweight tuning) that can further enhance performance in specific scenarios. These automated strategies are effective when the target is \emph{explicitly} specified (by instance identity or direct text grounding). Yet they are not designed to infer targets from \emph{implicit} instructions that require multi-step reasoning over spatial relations and geographic commonsense, precisely where a dedicated reasoning module becomes necessary to translate abstract semantics into actionable geometric prompts.

\subsection{Toward Reasoning-Driven Segmentation with LVLMs}

The landscape of vision-language understanding has shifted from simple perception to complex multi-modal reasoning. 
Modern LVLMs~\citep{liu2023improvedllava,bai2025qwen2,zhu2025internvl3} demonstrate remarkable capabilities in multi-step inference via Chain-of-Thought (CoT) strategies~\citep{wei2022chain,kojima2022large}.
Such capability is critical for Visual Question Answering (VQA) tasks involving spatial and causal logic, where models must deduce object relationships and functions beyond simple identification~\citep{hudson2019gqa,zellers2019recognition}.
Furthermore, recent studies have leveraged reinforcement learning to enhance the spatial reasoning abilities of LVLMs, leading to more precise grounding and deeper visual understanding~\citep{shen2025vlm,li2025videochat,zhang2025r1}. 
However, these approaches typically restrict their outputs to textual descriptions or coarse bounding boxes, lacking the granularity required for fine-grained shape analysis.

To extend this reasoning capacity to dense prediction, LISA \citep{lai2024lisa} establishes the embedding-as-mask paradigm by introducing a \texttt{<SEG>} token that allows an LVLM to interface with a mask decoder, sparking a line of work that explores how LVLMs can guide dense prediction. 
Along this line, LISA++~\citep{yang2023lisa++} extends from semantic to instance-level segmentation and enables more flexible text interaction, while GSVA~\citep{xia2024GSVA} augments the interface with a \texttt{<REJ>} token to handle empty-target cases.
Additionally, PixelLM~\citep{ren2024pixellm} introduces a multi-scale segmentation codebook and a lightweight pixel decoder, enabling the concurrent resolution of multiple open-set targets.
Departing from embedding-as-mask, Text4Seg~\citep{lan2024text4seg} reframes segmentation as text generation: the LVLM outputs coarse masks that are subsequently refined by SAM, avoiding additional decoder training. In contrast, SAM4MLLM~\citep{chen2024sam4mllm} fine-tunes the LVLM to directly produce geometric prompts for SAM, sidestepping specialized tokens and architectural changes.

Despite their architectural diversity, these methods predominantly rely on Supervised Fine-Tuning (SFT). While effective for instruction following, SFT treats the mapping from reasoning to spatial outputs as imitation learning via a next-token prediction loss. This paradigm tends to overfit specific training samples rather than learning generalizable prompting strategies, thereby limiting performance in open-ended scenarios~\citep{chu2025sft}.
Unlike SFT, Reinforcement Learning (RL) optimizes the model by aligning the model directly with the reward function, which allows the policy to explore and discover optimal strategies.
Among RL algorithms, standard algorithms like Proximal Policy Optimization (PPO)~\citep{schulman2017proximal} require a separate critic model, which is often prohibitive for large LVLMs. Alternatively, Direct Preference Optimization (DPO)~\citep{rafailov2023direct} necessitates high-quality paired preference data difficult to derive from binary masks. 
Consequently, Group Relative Policy Optimization (GRPO)~\citep{shao2024deepseekmath} emerges as the most suitable choice for the SAM prompt generation task. By estimating baselines from the average reward of a group of outputs, GRPO eliminates the need for a critic while efficiently optimizing the policy based on scalar rewards.
Adopting this paradigm, Seg-Zero~\citep{liu2025seg} and its successor VisionReasoner~\citep{liu2025visionreasoner} employ GRPO to train an LVLM to generate prompts for SAM2. However, their reward design depends on predefined box--point as intermediate pseudo ground truth, which implies costly instance-level annotations and narrows the search space—limitations that are particularly acute in remote sensing.

The LVLM-driven paradigms have begun to appear in the RS literature as well. For example, GeoPix~\citep{Ou2025geopix} adopts a \texttt{<SEG>}-style interface, while RSUniVLM~\citep{liu2024rsunivlm} and GeoGround~\citep{zhou2024geoground} follow a Text4Seg-like pipeline to produce masks from language. Yet, these methods focus on \emph{explicit} expressions and struggle with \emph{implicit, compositional} instructions that require multi-step geospatial reasoning. SegEarth-R1~\citep{li2025segearth} addresses this gap by formalizing \emph{geospatial pixel reasoning} (i.e., reasoning segmentation in RS) and releasing the EarthReason dataset. However, its architecture couples LVLM embeddings with a specialized mask generator, which reduces modularity, increases training overhead, and compresses reasoning into a single latent embedding, creating an information bottleneck that limits interpretability. RemoteReasoner~\citep{yao2025remotereasoner} further adopts a Seg-Zero–style GRPO framework for EarthReason, but still relies on mask-to-box conversion to provide pseudo box--point supervision, inheriting similar annotation and search-space constraints.

\textbf{Synthesis and Positioning.} Synthesizing the reviewed literature, effective RS reasoning segmentation demands a framework that bridges implicit linguistic reasoning with fine-grained visual execution, while robustly handling diverse annotation granularities. Concurrent with recent explorations such as Seg-Zero~\citep{liu2025seg}, VisionReasoner~\citep{liu2025visionreasoner}, and RemoteReasoner~\citep{yao2025remotereasoner}, which also adopt a decoupled LVLM–SAM paradigm, we develop Think2Seg-RS for RS reasoning segmentation. The crucial distinction lies in supervision: Seg-Zero and VisionReasoner rely on pseudo box--point labels generated from natural-image datasets, while RemoteReasoner applies mask-to-box conversion on EarthReason---all of them incur costly instance-level prompt supervision and constrain the search space. By contrast, Think2Seg-RS is optimized with GRPO using only mask annotations and a simple result-oriented reward (IoU + format). This mask-only optimization enlarges the exploration space for prompting, reduces annotation cost, and improves scalability to RS reasoning segmentation tasks.

\section{METHODOLOGY}

\begin{figure*}[!t]
    \centering
    \includegraphics[width=0.98\textwidth]{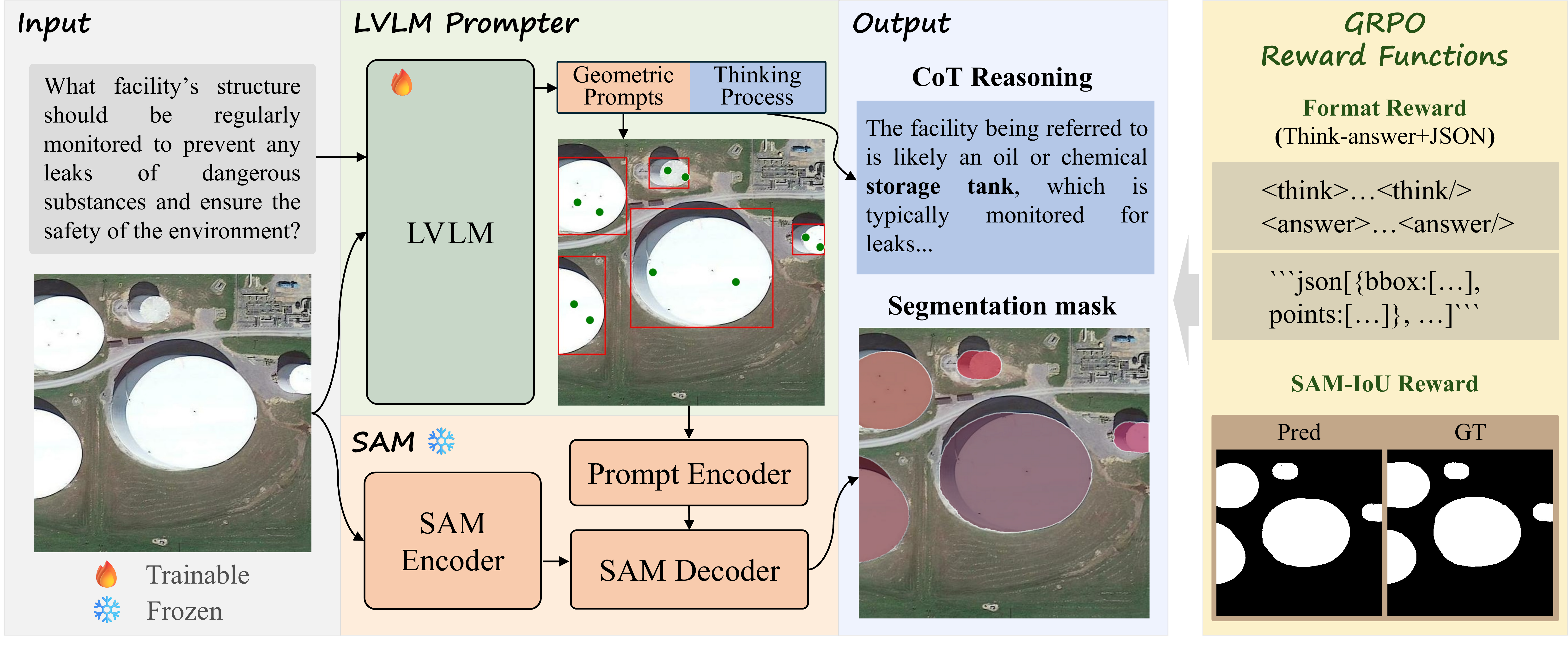}
    \caption{Architecture of Think2Seg-RS. A trainable LVLM prompter interprets the image–query pair, generates CoT reasoning, and outputs structured JSON prompts optimized via GRPO. These prompts are then executed by a frozen segmenter (SAM) to produce the final segmentation masks.}
    \label{fig:RSAM-Prompter framework}
\end{figure*}

\subsection{Think2Seg-RS Framework}

The architecture of Think2Seg-RS, illustrated in Fig.~\ref{fig:RSAM-Prompter framework}, follows a two-stage decoupled reasoning–execution design that combines a trainable LVLM prompter with a frozen segmentation engine. Specifically, we adopt Qwen-2.5-VL \citep{bai2025qwen2} as the LVLM prompter and SAM2 \citep{ravisam} as the universal segmenter. 
By explicitly separating high-level reasoning from low-level execution, this design reframes the task as a more tractable reasoning-to-prompt problem: the prompter learns to infer and generate high-quality geometric prompts that faithfully reflect user intent, while the segmenter converts these prompts into precise masks. This approach avoids costly fine-tuning of SAM, enhances modularity, and ensures scalability to future LVLMs, while the explicit CoT process equips the system to handle complex, implicit queries characteristic of RS imagery. 
Given a remote sensing image $I \in \mathbb{R}^{H \times W \times 3}$ and a complex natural language query $Q$, the framework proceeds in two stages.

\subsubsection{Reasoning and Prompting}
The LVLM prompter is the only trainable component of the framework and serves as the reasoning module. Upon receiving the input pair $(I,Q)$, together with an instruction template that guides its reasoning process (see Fig.~\ref{fig:prompt_template}), the prompter performs two tasks. 

First, it generates a Chain-of-Thought (CoT) enclosed in \verb|<think>| and \verb|</think>| tags, verbalizing its interpretation of the query in the context of the RS scene (e.g., ``The facility being referred to is likely an oil or chemical storage tank, which is typically monitored for leaks...''). This step enhances reasoning robustness and provides interpretability. 
Crucially, this step translates abstract query constraints into concrete visual definitions (e.g., shape, category) and explicit spatial logic. Since the model is autoregressive, these generated reasoning tokens serve as an explicit semantic anchor that materially conditions the subsequent prompting stage, thereby narrowing the visual search space and steering the geometric output away from semantic ambiguity.

Second, conditioned on this reasoning, the prompter produces structured geometric prompts for SAM2, denoted as \verb|P| in JSON format within \verb|<answer>| tags. Each such prompt specifies one bounding box and two positive points for a target instance, while an empty list \verb|[]| is returned when no target is present to preserve JSON validity. For subsequent formulation, we denote this prompt set abstractly as $P$, from which coordinate sets are derived and passed to the segmenter for execution.

\subsubsection{Segmentation Execution}
The prompts $P$ are parsed into coordinate sets $P_{\text{geom}}$, each corresponding to a target instance. For instance $i$, the SAM2 segmenter takes $(I, P_{\text{geom}}^{(i)})$ as input and produces a binary mask $M^{(i)} \in \{0,1\}^{H \times W}$. When multiple instances are present, the final mask is aggregated by logical union $M = \bigcup_{i=1}^{N} M^{(i)}$, where $N$ denotes the number of detected instances. This instance-aware execution ensures accurate delineation of each object while preserving spatial coherence across the entire scene.

\subsection{Mask-Only GRPO Optimization for Prompt Learning}

A central challenge in our framework lies in training the LVLM to generate effective prompts that faithfully bridge high-level reasoning with low-level segmentation. A naive zero-shot application of general-purpose LVLMs such as Qwen-2.5-VL~\citep{bai2025qwen2} or InternVL3~\citep{zhu2025internvl3} to RS imagery often produces inaccurate localization, with misaligned bounding boxes or poorly placed points, highlighting the need for robust task-specific adaptation.

A conventional supervised fine-tuning strategy, however, is ill-suited for this problem. For any given image $I$ and query $Q$, there exists a virtually infinite space of prompt \textit{combinations} (the number and types of prompts, e.g., boxes or points) and \textit{geometries} (the precise size and coordinates) that could yield correct segmentation. Consequently, there is no unique ground-truth prompt to serve as a fixed supervision target. Forcing the model to mimic a single pseudo prompt, as done in Seg-Zero~\citep{liu2025seg}, VisionReasoner~\citep{liu2025visionreasoner},  and RemoteReasoner~\citep{yao2025remotereasoner}, introduces behavioral cloning bias, restricts exploration to handcrafted or converted box--point annotations, and incurs annotation overhead that limits scalability in RS settings.

To overcome these limitations, we propose a mask-only reinforcement learning paradigm built on three design choices. First, we employ the GRPO algorithm~\citep{shao2024deepseekmath}, which efficiently leverages relative rewards among multiple candidates per query to stabilize large-model training. Second, we adopt a simple, result-oriented reward that directly evaluates the SAM2-produced mask: it consists of (i) a format reward ensuring valid \verb|<think>|/\verb|<answer>| tags and JSON outputs, and (ii) an IoU reward measuring overlap with the ground-truth mask. This direct feedback loop aligns prompt generation with final segmentation quality, avoiding the need for costly pseudo box--point supervision. Third, we fix the prompt combination to a universally effective set for RS targets (one bounding box and two positive points per instance), reducing the action space and making exploration tractable, while leaving the LVLM free to optimize spatial placement.

This mask-only GRPO optimization enables the LVLM to autonomously discover prompting strategies that generalize beyond handcrafted heuristics, enlarging the search space, reducing annotation cost, and enhancing scalability for RS reasoning segmentation.

\subsubsection{GRPO-based Prompt Optimization}

In our framework, the LVLM itself serves as the policy model $\pi_\theta$, responsible for generating structured JSON prompts given an image $I$ and query $Q$. To enable exploration, for each $(I, Q)$ pair we draw a set of $G$ candidate prompt outputs 
$O = \{o_1, o_2, \dots, o_G\}$, 
sampled from the old policy $\pi_{\theta_{\text{old}}}$ via stochastic decoding. Here, diverse candidates are drawn from the model’s output distribution for relative comparison. For each candidate $o_i \in O$, the frozen SAM2 performs segmentation conditioned on $o_i$, producing a mask from which a task-specific reward $r_i = R(o_i)$ is computed (detailed in Eq.~\ref{eq:reward}).

GRPO optimizes the LVLM by comparing the relative quality of candidates within each sampled group. Specifically, the rewards $\{r_1,\dots,r_G\}$ are normalized into standardized advantages:
\begin{equation}
a_i = \frac{r_i - \operatorname{mean}(\{r_j\}_{j=1}^G)}{\operatorname{std}(\{r_j\}_{j=1}^G)}.
\end{equation}
This allows the model to reinforce candidates that perform better than the group average, while discouraging weaker ones.

The policy parameters $\theta$ are updated by maximizing the GRPO objective:
\begin{equation}
\label{eq:GRPO}
\begin{split}
\mathcal{J}_{GRPO}(\theta) = \mathbb{E}_{O \sim \pi_{\theta_{\text{old}}}} \Bigg[ \frac{1}{G}\sum_{o_i \in O} \min \Bigg( 
&\frac{\pi_{\theta}(o_i)}{\pi_{\theta_{\text{old}}}(o_i)} a_i, \\
&\mathrm{clip}\!\left(\frac{\pi_{\theta}(o_i)}{\pi_{\theta_{\text{old}}}(o_i)}, 1-\epsilon, 1+\epsilon\right) a_i \Bigg) \\
& - \beta \,\mathrm{D}_{KL}(\pi_{\theta} \parallel \pi_{ref}) \Bigg],
\end{split}
\end{equation}
where the clipping stabilizes training by limiting the update ratio to a range defined by $\epsilon$, and the Kullback–Leibler (KL) divergence penalty term $\mathrm{D}_{KL}(\pi_{\theta}\parallel \pi_{ref})$ (weighted by $\beta$) prevents the LVLM from deviating excessively from the reference model $\pi_{ref}$ (typically the initialization checkpoint).  

In this way, GRPO directly aligns the LVLM’s prompt generation policy with downstream segmentation performance: the LVLM progressively learns to generate bounding boxes and points that maximize IoU after execution by SAM2, without requiring any handcrafted prompt supervision.

\subsubsection{Result-Oriented Reward Function}

Our reward function provides direct supervision from the final segmentation result. For each output $o_i$, the total reward is defined as:
\begin{equation}
\label{eq:reward}
R(o_i) = \lambda_1 R_{format}(o_i) + \lambda_2 R_{IoU}(o_i),
\end{equation}
where $\lambda_1$ and $\lambda_2$ balance the contributions of the two components.  

Format reward $R_{format}$ ensures the output is syntactically valid: it checks both the correct use of the \texttt{<think>}/\texttt{<answer>} tags and the validity of the JSON object. This guarantees that the generated prompts can always be parsed by the segmentation engine.  

Segmentation reward $R_{IoU}$ measures the IoU between the predicted mask $M_{pred}$ (produced by SAM2) and the ground-truth mask $M_{gt}$. In empty-target scenarios, $R_{IoU}$ is set to 1 if the model outputs an empty list and 0 otherwise. This direct IoU feedback aligns prompt generation with final segmentation performance, reinforcing strategies that improve execution outcomes.

\subsubsection{Prompting Strategy for Tractability}

Exploring both prompt combinations (types and numbers) and geometries (positions and sizes) would create an intractably large action space. To simplify the learning problem while preserving generality, we constrain the prompt set to a universally effective configuration for RS: one bounding box and two positive points per instance. This design allows the LVLM to focus its optimization on spatial placement and scale, which are the most critical factors for accurate segmentation. By fixing the combination and learning only the geometry, GRPO training remains efficient and stable while still enabling diverse prompting strategies to emerge.

\section{Experiments and Results}

\subsection{Experimental Settings}
The experimental setup is described in three parts, covering datasets, evaluation metrics, and implementation details to ensure reproducibility and fair comparison.

\subsubsection{Datasets}
The reasoning segmentation task is evaluated on the EarthReason dataset~\citep{li2025segearth}, currently the only publicly available benchmark for optical remote sensing reasoning segmentation.  
EarthReason contains 5,434 images across 28 categories, with the official split consisting of 2,731 training, 1,135 validation, and 1,928 test samples.  
These subsets include 60, 20, and 39 empty-target cases, respectively, which are critical for evaluating rejection capability.  
The provided ground-truth masks served as supervision during training and evaluation. 

To further examine whether a model trained for reasoning segmentation can generalize to explicit referential grounding, a zero-shot transfer evaluation was performed.  
Specifically, the models trained solely on the EarthReason training set were directly applied to the test sets of three referring segmentation benchmarks, including RRSIS-D~\citep{liu2024rrsis-d}, RISBench~\citep{dong2024cross}, and RefSegRS~\citep{rrsis}, without any additional fine-tuning.  
The RRSIS-D test set contains 3,481 samples across 20 categories, the RISBench test set includes 16,158 samples covering 26 categories, and the RefSegRS test set comprises 1,817 samples across 20 categories.

It is important to clarify the difference in annotation granularity between reasoning and referring segmentation datasets.  
The EarthReason dataset adopts \emph{semantic-level masks}, in contrast to the instance-level masks used in most referring segmentation benchmarks.  
A semantic-level mask corresponds to a semantically coherent concept such as ``greenhouse'' or ``residential area,'' which may include multiple spatially disjoint regions that share the same functional or visual meaning.  
Unlike instance-level annotations that delineate individual objects, semantic-level masks emphasize semantic consistency rather than spatial continuity, aligning with the nature of reasoning segmentation where the target concept is often defined by function or context rather than by a single geometric instance.

\subsubsection{Evaluation Metrics}
Following prior works \citep{kazemzadeh2014referitgame,yu2016modeling,lai2024lisa}, segmentation performance was measured using generalized Intersection-over-Union (gIoU) and cumulative Intersection-over-Union (cIoU).  
For a predicted mask $M_{pred}$ and ground-truth mask $M_{gt}$, with $N$ denoting the total number of samples, gIoU is defined as:
\begin{equation}
gIoU = \frac{1}{N}\sum_{i=1}^{N}\frac{|M_{pred} \cap M_{gt}|}{|M_{pred} \cup M_{gt}|},
\end{equation}
and cIoU is defined as:
\begin{equation}
cIoU = \frac{\sum_{i=1}^{N}|M_{pred} \cap M_{gt}|}{\sum_{i=1}^{N}|M_{pred} \cup M_{gt}|}.
\end{equation}
The gIoU reflects per-sample segmentation quality, while cIoU aggregates intersection-over-union performance across the dataset.

For the referring segmentation experiments, we additionally report P@0.5, which measures the proportion of test samples whose IoU between the predicted mask and the ground truth exceeds a threshold of 0.5:
\begin{equation}
P@0.5 = \frac{1}{N} \sum_{i=1}^{N} \mathbb{I}\left( \frac{|M_{\text{pred}} \cap M_{\text{gt}}|}{|M_{\text{pred}} \cup M_{\text{gt}}|} > 0.5 \right),
\end{equation}
where $\mathbb{I}(\cdot)$ is the indicator function.  
This metric reflects the model’s ability to produce accurate and confident segmentations that meet a minimum overlap threshold with the ground-truth region.

\subsubsection{Implementation Details}
The framework was implemented in PyTorch with DeepSpeed \citep{rasley2020deepspeed}. Qwen-2.5-VL-3B \citep{bai2025qwen2} was adopted as the default LVLM prompter, and a frozen SAM2-Small \citep{ravisam} was used as the default segmentation module. All models were trained with bf16 precision. Input images were resized to $840 \times 840$ pixels to match the architectural requirement of Qwen-2.5-VL’s vision encoder, which processes images in $14 \times 14$ patches and requires dimensions divisible by 28.  

Optimization employed AdamW \citep{loshchilov2017decoupled} with a learning rate of $1 \times 10^{-6}$, no weight decay, and a total batch size of 16. The hyperparameters in Eq.~\ref{eq:reward} were set to $\lambda_1=1$ and $\lambda_2=2$, emphasizing the segmentation reward as the dominant signal for policy optimization. For GRPO training, the number of generations per sample was set to $G=16$ to encourage exploration of the prompt space, and the KL divergence coefficient ($\beta$ in Eq.~\ref{eq:GRPO}) was set to $10^{-3}$. All experiments were conducted on four NVIDIA A800 GPUs (80 GB memory each). The implementation was built upon the open-source VLM-R1 framework \citep{shen2025vlm}.

\begin{table*}[!htb]
    \caption{Comparison of reasoning segmentation results on the EarthReason dataset}
    \label{tab:earthreason_results}
    \centering
    \small
    \begin{threeparttable}
        \begin{tabular*}{1.0\textwidth}{@{\extracolsep{\fill}}llcccc}
            \toprule
            \multirow{2}{*}{Method} & \multirow{2}{*}{Model} & \multicolumn{2}{c}{Validation} & \multicolumn{2}{c}{Test} \\
            \cmidrule(lr){3-4} \cmidrule(lr){5-6}
             &  & gIoU & cIoU & gIoU & cIoU \\
            \midrule
            \rowcolor[HTML]{F2F2F2}
            \multicolumn{6}{@{}l@{}}{\textit{End-to-End Reasoning–Segmentation Models}} \\
            LISA$^*$~\citep{lai2024lisa}    & CLIP-L (304M) + Vicuna-7B   & 61.04 & 57.39 & 60.88 & 59.10 \\
            PixelLM$^*$~\citep{ren2024pixellm}  & CLIP-L (304M) + Vicuna-7B   & 57.94 & 57.79 & 60.01 & 59.22 \\
            PSALM$^*$~\citep{zhang2024psalm}    & Swin-B (88M) + Phi-1.5-1.3B & 66.61 & 62.03 & 68.30 & 64.61 \\
            SegEarth-R1~\citep{li2025segearth}  & Swin-B (88M) + Phi-1.5-1.3B & 68.60 & 64.13 & 70.75 & 68.25 \\
            
            \rowcolor[HTML]{F2F2F2}
            \multicolumn{6}{@{}l@{}}{\textit{Decoupled LVLM–SAM Frameworks}} \\
            \textcolor{gray}{Pseudo GT (Seg-Zero)}      & \textcolor{gray}{-- + SAM2-Large}                 & \textcolor{gray}{81.30} & \textcolor{gray}{80.18} & \textcolor{gray}{82.16} & \textcolor{gray}{78.40} \\
            \textcolor{gray}{Pseudo GT (VisionReasoner)} & \textcolor{gray}{-- + SAM2-Large}                 & \textcolor{gray}{84.65} & \textcolor{gray}{84.85} & \textcolor{gray}{85.57} & \textcolor{gray}{85.27} \\

            RemoteReasoner~\citep{yao2025remotereasoner} & Qwen-2.5-VL-7B + SAM2$^\ddag$ & 69.02 & 67.80 & 70.96 & 69.13 \\
            Seg-Zero$^\dag$~\citep{liu2025seg}  & Qwen-2.5-VL-7B + SAM2-Large & 67.15 $\pm$ 0.44 & 65.41 $\pm$ 0.64 & 67.89 $\pm$ 0.42 & 62.43 $\pm$ 3.50 \\
            VisionReasoner$^\dag$~\citep{liu2025visionreasoner}  & Qwen-2.5-VL-7B + SAM2-Large & 66.58 $\pm$ 0.24 & 67.13 $\pm$ 0.81 & 67.19 $\pm$ 0.22 & 65.41 $\pm$ 0.81 \\
            Think2Seg-RS (ours)  & Qwen-2.5-VL-3B + SAM2-Small & 69.23 $\pm$ 0.77 & 71.30 $\pm$ 1.20 & 70.81 $\pm$ 0.25 & 74.21 $\pm$ 0.81 \\
            Think2Seg-RS (ours)  & Qwen-2.5-VL-7B + SAM2-Small & \textbf{72.16 $\pm$ 0.69} & \textbf{72.87 $\pm$ 0.88} & \textbf{73.36 $\pm$ 0.68} & \textbf{75.60 $\pm$ 0.64} \\
            \bottomrule
        \end{tabular*}
        \begin{tablenotes}
            \item \textit{Note:} 
            For methods evaluated by us (including our proposed models and baselines marked with $^\dag$), results are reported as mean $\pm$ standard deviation.
            $^*$ denotes results reproduced by the SegEarth-R1 paper~\citep{li2025segearth}; 
            $^\dag$ indicates results reproduced by us on the EarthReason dataset; 
            $^\ddag$ indicates that the SAM2 version was not specified in the RemoteReasoner paper~\citep{yao2025remotereasoner}.
            The \textcolor{gray}{gray} entries represent the pseudo ground-truth prompt baselines where geometric prompts are analytically derived from ground-truth masks and fed directly to SAM2, providing a performance upper bound for the pseudo ground-truth supervision methods.
        \end{tablenotes}
    \end{threeparttable}
\end{table*}

\subsection{Reasoning Segmentation Results}

The main results on the EarthReason dataset~\citep{li2025segearth} are summarized in Table~\ref{tab:earthreason_results}. Following the experimental protocol of SegEarth-R1~\citep{li2025segearth}, the reference model introduced alongside EarthReason, all methods were trained exclusively on the official training split to ensure fair comparison.

For clarity, we categorize existing approaches into two main families.  
(1) End-to-end reasoning--segmentation models jointly optimize reasoning and segmentation within a unified architecture by fine-tuning the mask decoder together with the vision–language backbone (e.g., LISA~\citep{lai2024lisa}, PixelLM~\citep{ren2024pixellm}, PSALM~\citep{zhang2024psalm}, SegEarth-R1~\citep{li2025segearth}).  
(2) Decoupled LVLM–SAM frameworks, which employ an LVLM to guide a frozen segmenter via geometric prompts (e.g., Seg-Zero~\citep{liu2025seg}, VisionReasoner~\citep{liu2025visionreasoner}, RemoteReasoner~\citep{yao2025remotereasoner}, and our proposed Think2Seg-RS).

To evaluate Seg-Zero~\citep{liu2025seg} and its successor VisionReasoner~\citep{liu2025visionreasoner} on remote sensing imagery, we adapted their training pipelines to EarthReason. Both methods were originally designed for natural image datasets such as RefCOCOg~\citep{yu2016modeling}, which provide instance-level masks enabling direct generation of pseudo box–point supervision. In contrast, EarthReason supplies only a single semantic-level binary mask for each image, sometimes covering several spatially distinct objects (e.g., clustered storage tanks in Fig.~\ref{fig:EarthReason_results}).  
To reproduce Seg-Zero, we followed its original approach by deriving one tight bounding box and two interior points from the entire mask. For VisionReasoner, we first decomposed each semantic-level mask into sub-instances by clustering foreground pixels using DBSCAN~\citep{ester1996density}, then generated a separate box–point pair for each cluster. Both reproductions were trained and evaluated under identical settings.

\subsubsection{End-to-End vs.\ Decoupled}

Across both validation and test splits, decoupled frameworks consistently outperform end-to-end counterparts. Although the mask decoder remains frozen, the LVLM–SAM family (including RemoteReasoner and our model) benefits from three key factors:  
(i) SAM2’s strong zero-shot segmentation prior learned from large-scale pretraining,  
(ii) a clean division between high-level language reasoning and low-level pixel execution, and  
(iii) reinforcement learning that directly ties prompt quality to downstream segmentation accuracy.  
In contrast, end-to-end models must align compressed LVLM embeddings with a learnable decoder, which often creates optimization bottlenecks and limits generalization when facing open-ended, free-form natural language queries.

\subsubsection{Against End-to-End SFT Methods}

Relative to the strongest end-to-end baseline (SegEarth-R1), Think2Seg-RS (7B) achieves mean improvements of +2.61 gIoU and +7.35 cIoU on the test set. The lighter 3B variant yields comparable gIoU but maintains a substantial advantage in cIoU (+5.96), demonstrating that a reinforcement learning–optimized prompting policy is more scalable than training a task-specific decoder. Notably, the cIoU improvement significantly exceeds the gIoU gain, indicating superior global completeness across the dataset. Unlike SFT models limited by decoder bottlenecks, Think2Seg-RS leverages the robust spatial priors of the frozen SAM and optimizes directly against the final mask quality, allowing the LVLM to discover flexible prompting patterns that maximize holistic object coverage.

\subsubsection{Comparison Within Decoupled Methods}

Within the decoupled family, we compare four representative frameworks: Seg-Zero, VisionReasoner, RemoteReasoner, and our Think2Seg-RS.  
As shown in Table~\ref{tab:earthreason_results}, RemoteReasoner~\citep{yao2025remotereasoner} is the strongest published baseline. Evaluated across four independent runs, our Think2Seg-RS (7B) further pushes the state of the art to $73.36 \pm 0.68$ gIoU and $75.60 \pm 0.64$ cIoU on the test set, yielding mean improvements of +2.40 and +6.47 points over RemoteReasoner. Confidence interval analysis confirms these gains are statistically significant: the 95\% confidence interval for our 7B model's gIoU ($[72.28, 74.44]$) on the test set establishes a lower bound that reliably exceeds the RemoteReasoner baseline ($70.96$). The smaller 3B model also performs competitively ($70.81 \pm 0.25$ gIoU, $74.21 \pm 0.81$ cIoU), remaining comparable to RemoteReasoner in gIoU and clearly superior in cIoU (+5.08).

The pronounced cIoU gain suggests that Think2Seg-RS produces more globally coherent and complete segmentation masks, as the mask-level reinforcement signal encourages the model to group spatially disjoint but semantically related components into a unified prediction. While gIoU primarily captures per-instance geometric accuracy, cIoU reflects overall coverage and cross-scene consistency, highlighting that our mask-based optimization improves both spatial completeness and prediction stability across diverse samples.

The performance gap among decoupled methods can be largely attributed to differences in supervision and reward design. 
Both Seg-Zero and VisionReasoner rely on pseudo instance-level box–point supervision, which is unavailable in EarthReason. 
To reproduce these methods, we derived one bounding box and two points from each semantic-level mask for Seg-Zero, and applied DBSCAN to decompose masks into sub-instances before generating box–point pairs for VisionReasoner. 
However, both strategies inevitably introduce noisy geometric supervision due to the lack of true instance annotations, forcing the LVLM to mimic heuristic labels rather than optimizing for segmentation fidelity, leading to degraded performance. Notably, Seg-Zero exhibits severe cIoU instability ($\pm 3.50$ on the test set), as encompassing disjoint targets within a single global box triggers inconsistent background over-segmentation driven by minor variations in point placement.
RemoteReasoner extends this idea to remote sensing by converting masks into boxes through a mask-to-box procedure and adopting a composite reward that balances bounding box IoU, count, and format validity. This design alleviates annotation noise and aligns policy learning with object separation and structural correctness, explaining its superior performance relative to our Seg-Zero and VisionReasoner reproductions.

Crucially, the pseudo ground-truth baselines in Table~\ref{tab:earthreason_results} expose a fundamental limitation of the pseudo-supervision paradigm. 
While the SAM2 executor demonstrates a theoretical ceiling exceeding 80\% cIoU, the supervised counterparts (Seg-Zero and VisionReasoner) lag behind by nearly 20 points, revealing that the bottleneck lies in prompt quality rather than segmentation capacity. 
This execution gap arises because these methods tether optimization to heuristic labels rather than the ultimate segmentation goal. 
In the context of RS imagery, where targets exhibit complex or ambiguous morphologies, such rigid geometric constraints often serve as suboptimal proxies for the ideal prompt. 
Consequently, the policy is penalized for deviating from these heuristics even if doing so would trigger superior mask generation, creating a misalignment between the proxy reward (fitting the pseudo ground-truth) and the true objective (accurate segmentation).

In contrast, Think2Seg-RS eliminates the need for any pseudo instance supervision. The LVLM is optimized directly through mask-level reinforcement using only the final segmentation quality as feedback. This mask-only optimization avoids the bias introduced by synthetic boxes and points and allows the model to explore a wider range of prompting strategies. Consequently, Think2Seg-RS achieves higher scene-level completeness and more accurate boundary delineation, as reflected in its substantial cIoU improvement.

\begin{figure*}[!t]
\centering
\includegraphics[width=0.95\textwidth]{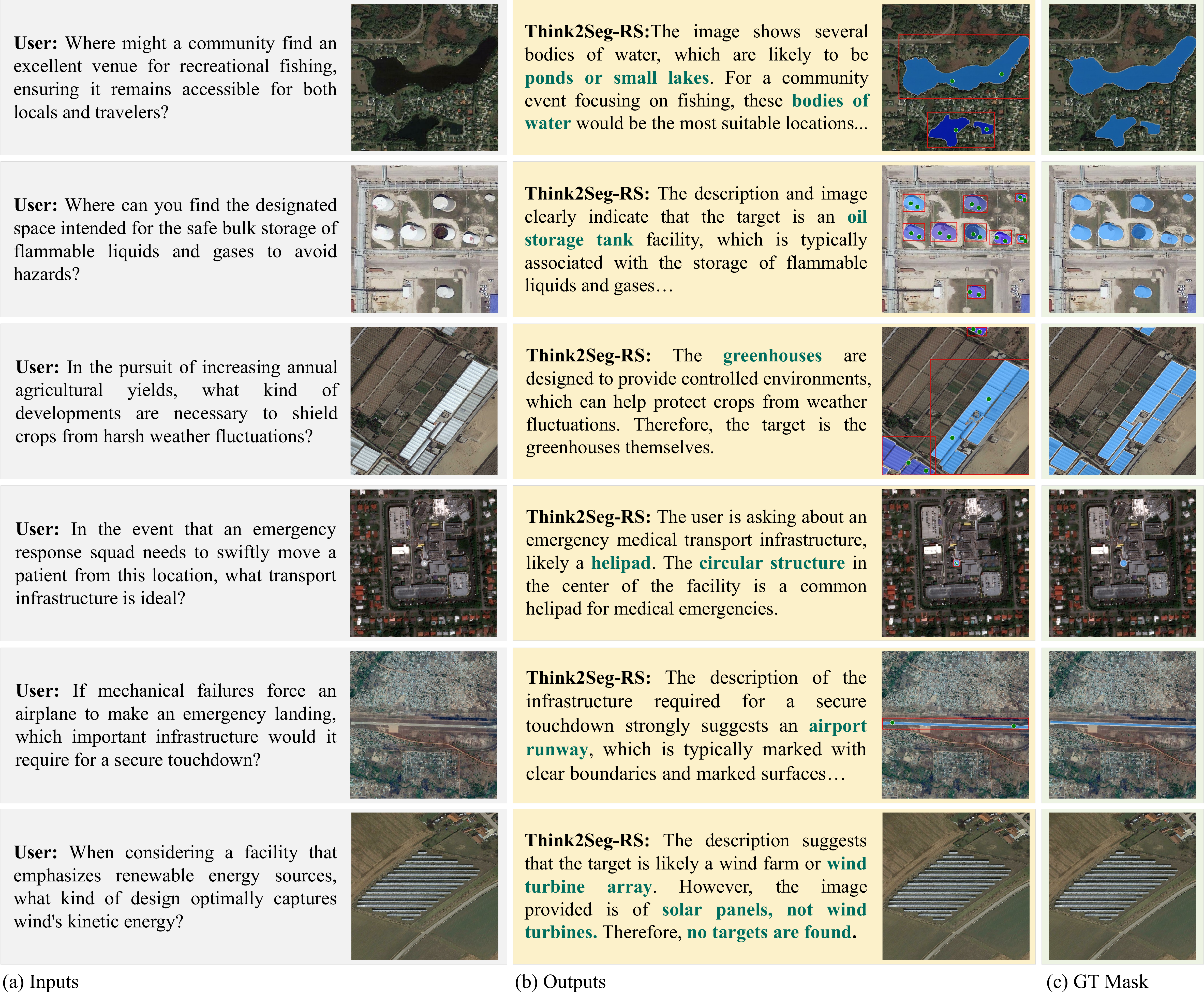}
\caption{Qualitative results of Think2Seg-RS on the EarthReason dataset. 
(a) Inputs, including the user query and corresponding remote sensing image. 
(b) Model outputs, comprising the LVLM's reasoning text from the \texttt{<think>} stage, the generated geometric prompts (bounding boxes in red and positive points in green), and the resulting segmentation mask predicted by SAM2. 
(c) Ground-truth (GT) mask for reference.}
\label{fig:EarthReason_results}
\end{figure*}

\subsubsection{Qualitative Behavior}

Figure~\ref{fig:EarthReason_results} presents representative examples illustrating the reasoning traces and geometric prompting behaviors of Think2Seg-RS across diverse remote sensing scenarios.  
In each case, the model first verbalizes its reasoning process in the \verb|<think>| stage.
Crucially, this stage functions as a semantic-to-geometric bridge, it interprets the query to define specific visual constraints (e.g., morphology, category, and exclusion criteria), which then serve as the conditioning context for generating structured prompts in the \verb|<answer>| stage.

In the first example, responding to a query about a recreational fishing venue, the model correctly identifies the target as ``bodies of water''.
This semantic translation materially affects the output by restricting the visual search space to lake textures.
It then generates precise box and points for the single large waterbody, and an encompassing box that groups the two smaller, adjacent waterbodies, each marked by a distinct point, showcasing its flexible prompts for SAM.

The second example, depicting an oil storage facility, highlights the model’s handling of multi-instance regions. Its reasoning connects ``storage of flammable liquids'' with ``oil storage tanks,'' producing distinct boxes for each tank even though the ground-truth annotation merges them into one mask.  

In the third row, featuring a greenhouse complex, Think2Seg-RS reasons that ``the greenhouses are designed to provide controlled environments'' and demonstrates a flexible grouping strategy. It correctly recognizes that the target consists of spatially separate clusters and, accordingly, partitions them into three subgroups for segmentation. This reveals its ability to adapt its prompting strategy to the target’s spatial distribution.  

The fourth example showcases robust localization of a small yet semantically critical object in a dense urban environment. The model reasons that a ``helipad'' is a circular structure for medical emergencies, which serves as a morphological constraint to pinpoint the specific circular geometry amidst complex urban structures, and generates a tightly fitted prompt around the correct feature.  

In the fifth example, the model correctly interprets ``infrastructure required for a secure touchdown'' as an airport runway and generates elongated, axis-aligned prompts that precisely align with the target’s linear structure.

Finally, the last example illustrates reliable negative reasoning. The model correctly rejects an invalid instruction, reasoning that ``the image provided is of solar panels, not wind turbines.''
The conclusion ``no targets are found'' directly dictates the generation of an empty prompt list, showcasing its ability to recognize semantic mismatches and abstain from false predictions.
This rejection behavior will be further analyzed in Subsection~\ref{sec:rejection}.

Overall, these qualitative examples demonstrate that our Think2Seg-RS integrates interpretable reasoning with precise spatial execution. The textual reasoning traces provide insight into the decision process, while the geometric prompts reveal structured spatial understanding, together forming a coherent reasoning-to-execution pipeline capable of handling diverse structures, implicit semantics, and open-ended instructions in remote sensing imagery.

For a more comprehensive evaluation, the appendix presents additional examples (Fig.~\ref{fig:EarthReason_results_appendix}) and representative failure cases (Fig.~\ref{fig:failure_case}) of Think2Seg-RS on the EarthReason dataset.

\begin{table*}[htbp]
    \centering
    \small
    \begin{threeparttable}
    \caption{Detailed analysis of rejection capability on the EarthReason dataset.}
    \label{tab:rejection_analysis}
    \setlength{\tabcolsep}{12pt} 
    % \begin{tabular}{lccccc}
    \begin{tabular*}{1.0\textwidth}{@{\extracolsep{\fill}}lccccc}
        \toprule
        \multirow{2}{*}{Method} & \multicolumn{2}{c}{Predictions on Empty Scenes} & Errors on Target Scenes & \multicolumn{2}{c}{Quantitative Evaluation} \\
        \cmidrule(lr){2-3} \cmidrule(lr){4-4} \cmidrule(lr){5-6}
         & \begin{tabular}{@{}c@{}}Correct Rejection \\ (TP) $\uparrow$\end{tabular} & \begin{tabular}{@{}c@{}}Missed Rejection \\ (FN) $\downarrow$\end{tabular} & \begin{tabular}{@{}c@{}}Erroneous Rejection \\ (FP) $\downarrow$\end{tabular} & \begin{tabular}{@{}c@{}}Precision \\ (\%) $\uparrow$\end{tabular} & \begin{tabular}{@{}c@{}}Recall \\ (\%) $\uparrow$\end{tabular} \\
        \midrule
        \multicolumn{6}{@{}l}{\textit{Validation Set (20 Empty Samples)}} \\
        SegEarth-R1     & 7  & 13 & 49 & 12.5 & 35.0 \\
        Think2Seg-RS-3B & 9  & 11 & \textbf{4}  & \textbf{69.2} & 45.0 \\
        Think2Seg-RS-7B & \textbf{12} & \textbf{8}  & 10 & 54.5 & \textbf{60.0} \\
        \midrule
        \multicolumn{6}{@{}l}{\textit{Test Set (39 Empty Samples)}} \\
        SegEarth-R1     & 12 & 27 & 76 & 13.6 & 30.8 \\
        Think2Seg-RS-3B & 19 & 20 & \textbf{2}  & \textbf{90.5} & 48.7 \\
        Think2Seg-RS-7B & \textbf{24} & \textbf{15} & 7  & 77.4 & \textbf{61.5} \\
        \bottomrule
    \end{tabular*}
    \begin{tablenotes}
    \footnotesize
    \item \textit{Note:} Empty scenes are treated as the positive class. $\uparrow$ ($\downarrow$) indicates higher (lower) is better. TP (True Positive): Correct rejection of empty scenes (Correct Rejection); FN (False Negative): Failure to identify empty scenes (Missed Rejection); FP (False Positive): Erroneous rejection of target scenes (Erroneous Rejection).
    \end{tablenotes}
    \end{threeparttable}
\end{table*}

\subsection{Analysis of Rejection Capability}
\label{sec:rejection}

RS imagery often features complex, cluttered backgrounds and large intra-class variance, making it common for the queried object to be absent from a given scene. An effective reasoning–segmentation model must therefore recognize such absence and avoid hallucinating false positives, which can lead to misleading downstream analysis. To quantitatively assess this ability, we evaluated the rejection performance of Think2Seg-RS on the validation and test sets of the EarthReason dataset. Since the Seg-Zero framework lacks an inherent rejection mechanism, we compare our Think2Seg-RS (3B and 7B) with SegEarth-R1, which includes this capability.

As detailed in Table \ref{tab:rejection_analysis}, our framework demonstrates a distinct advantage in distinguishing empty targets. The baseline SegEarth-R1 exhibits high error rates in both directions, suffering from low recall on empty scenes and a high rate of erroneous rejections on valid targets (Precision: 13.6\%). This unreliable discrimination capability stems from its coupled architecture, which lacks an explicit mechanism to suppress uncertain predictions in cluttered backgrounds. In contrast, Think2Seg-RS significantly mitigates these errors. The 3B variant adopts a conservative policy, minimizing false rejections to achieve an exceptional precision of 90.5\%, whereas the 7B variant demonstrates enhanced sensitivity to absence cues, improving recall to 61.5\%. This scaling trend indicates that the larger model moves beyond the conservative rejection of simple cases to identify more subtle or ambiguous negative samples, albeit with a slight trade-off in precision. This robustness is fundamentally driven by our result-oriented GRPO optimization, which penalizes hallucinations through zero-IoU feedback, combined with reasoning process that allows the LVLM to verify semantic consistency before issuing geometric prompts. As illustrated in the last row of Figure \ref{fig:EarthReason_results}, this capability is substantiated by explicit reasoning (e.g., “no targets are found”), ensuring interpretable feedback rather than silent failure.

\begin{table*}[!htb]
    \caption{Zero-shot evaluation results on the referring segmentation benchmarks RRSIS-D, RISBench, and RefSegRS}
    \label{tab:rrsis_zero-shot_results}
    \centering
    \small
    \resizebox{\textwidth}{!}{
    \begin{threeparttable}
    \setlength{\tabcolsep}{4pt}
    % \begin{tabular*}{1.0\textwidth}{@{\extracolsep{\fill}}lccccccccc}
    \begin{tabular}{lccccccccc}
        \toprule
        \multirow{2}{*}{Method} & \multicolumn{3}{c}{RRSIS-D}                      & \multicolumn{3}{c}{RISBench}       & \multicolumn{3}{c}{RefSegRS}              \\
        \cmidrule(lr){2-4} \cmidrule(lr){5-7} \cmidrule(lr){8-10}
                                & gIoU           & cIoU           & P@0.5          & gIoU           & cIoU          & P@0.5          & gIoU           & cIoU          & P@0.5 \\
        \midrule
        Qwen-2.5-VL-3B+SAM2      & 14.87          & 15.86          & 12.94          & 15.85          & 12.80          & 15.58          & 4.18  & 6.23 & 0.74  \\
        Qwen-2.5-VL-7B+SAM2      & 28.14          & 30.45          & 37.06          & 26.30           & 25.76          & 31.15          & 5.16  & 8.29 & 3.09        \\
        SegEarth-R1             & 27.91         & 26.21           & 26.26         & 19.01          & 17.65          & 16.72    & 1.92  & 3.98  & 0.06    \\
        RemoteReasoner  & \textbf{50.97}         & -              & \textbf{54.29}         & - & - & - & - & - & -        \\
        Seg-Zero$^*$               & 44.63 $\pm$ 1.52 & 45.09 $\pm$ 1.09 & 46.81 $\pm$ 1.79 & 43.52 $\pm$ 1.61 & 35.83 $\pm$ 1.25 & 43.94 $\pm$ 1.89 & 10.14 $\pm$ 0.64 & 12.18 $\pm$ 1.35 & 4.14 $\pm$ 0.44 \\
        VisionReasoner$^*$ & 48.27 $\pm$ 0.90 & 52.67 $\pm$ 1.58 & 52.18 $\pm$ 1.24 & \textbf{50.21 $\pm$ 1.64} & \textbf{44.48 $\pm$ 2.26} & \textbf{53.36 $\pm$ 1.80}   & 11.07 $\pm$ 0.28 & 12.13 $\pm$ 0.66 & 4.58 $\pm$ 0.20       \\
        Think2Seg-RS-3B (ours)  & 42.21 $\pm$ 1.69 & 47.21 $\pm$ 2.68 & 42.92 $\pm$ 2.32 & 38.44 $\pm$ 1.47 & 34.84 $\pm$ 2.15 & 37.11 $\pm$ 2.04 & 14.64 $\pm$ 1.36 & 16.63 $\pm$ 2.15 & 5.92 $\pm$ 0.97     \\
        Think2Seg-RS-7B (ours)  & 46.40 $\pm$ 0.74 & \textbf{53.07 $\pm$ 0.84} & 49.01 $\pm$ 1.81 & 45.28 $\pm$ 1.46 & 43.99 $\pm$ 0.99 & 46.15 $\pm$ 2.78 & \textbf{15.13 $\pm$ 0.81} & \textbf{17.82 $\pm$ 1.04} & \textbf{6.32 $\pm$ 0.94} \\
        \bottomrule
    \end{tabular}
    \begin{tablenotes}
        \item \textit{Note:} For our models and baselines marked with $^*$, results are reported as mean $\pm$ standard deviation to verify statistical significance.
    \end{tablenotes}
    \end{threeparttable}
    }
\end{table*}

\subsection{Generalization on Referring Expression Segmentation}
\label{sec:transfer}

Reasoning segmentation and referring expression segmentation are closely related but differ fundamentally in task formulation.  
The former requires implicit and compositional reasoning that links textual descriptions to semantic regions through multi-step inference, whereas the latter involves explicit visual grounding of a directly mentioned target.  
Intuitively, a model capable of performing reasoning segmentation should, in principle, possess the necessary understanding to handle the simpler referring task.  
To verify this assumption and examine whether reasoning-trained models can generalize to explicit referential grounding, we conducted a challenging zero-shot transfer experiment.  

Specifically, the models optimized for the reasoning segmentation task on the EarthReason dataset~\citep{li2025segearth} were subsequently tested on three referring segmentation benchmarks, including RRSIS-D~\citep{liu2024rrsis-d}, RISBench~\citep{dong2024cross}, and RefSegRS~\citep{rrsis}, to evaluate their zero-shot transfer capability without any additional fine-tuning.
This setting jointly tests cross-dataset transfer (different visual distributions) and cross-task transfer (from reasoning to referring), providing a rigorous assessment of the model’s reasoning-to-referring adaptability.

\subsubsection{Results and Observations}

The results, summarized in Table~\ref{tab:rrsis_zero-shot_results}, present a comprehensive comparison across multiple model paradigms.  
To establish reference points, we first include two direct zero-shot baselines: Qwen-2.5-VL-3B+SAM2 and Qwen-2.5-VL-7B+SAM2, which use the off-the-shelf LVLMs and SAM executor without any training on EarthReason or the referring datasets.  
These models demonstrate the inherent capability of general-purpose LVLMs to perform coarse referring segmentation, yet their performance remains modest across all datasets, confirming that task-specific adaptation is essential for robust grounding in the RS domain.

Next, we evaluate SegEarth-R1~\citep{li2025segearth}, which represents an end-to-end SFT paradigm that jointly fine-tunes the reasoning and segmentation components. Although effective on the training dataset (see Table~\ref{tab:earthreason_results}), SegEarth-R1 performs poorly under zero-shot transfer. This sharp degradation highlights the limited generalizability of SFT-based models, which tend to overfit to dataset-specific visual--text priors and struggle to adapt when the data distribution or task formulation shifts.

Among the decoupled frameworks, Seg-Zero~\citep{liu2025seg} and VisionReasoner~\citep{liu2025visionreasoner}, both reproduced using our EarthReason-trained LVLM–SAM pipeline, show considerably higher transferability. Their instance-level box–point supervision enables more localized and fine-grained grounding, yielding superior performance on RRSIS-D and RISBench. The RemoteReasoner~\citep{yao2025remotereasoner} results, reported from its original paper, further improve upon these baselines via a refined mask-to-box generation and composite reward scheme, achieving $50.97$ gIoU and $54.29$ P@0.5 on RRSIS-D.

In contrast, our Think2Seg-RS models exhibit a markedly different behavior. While demonstrating strong semantic-level reasoning capability, they underperform instance-oriented methods on RRSIS-D and RISBench. However, on the RefSegRS benchmark, 
Think2Seg-RS (7B) attains the best overall performance among all models, yielding $15.13 \pm 0.81$ gIoU, $17.82 \pm 1.04$ cIoU, and $6.32 \pm 0.94$ P@0.5. Statistical significance is confirmed by non-overlapping 95\% confidence intervals (e.g., [16.16, 19.48] for our cIoU versus [11.08, 13.18] for VisionReasoner).

\subsubsection{Analysis and Interpretation}

The cross-task generalization behavior can be explained by the intrinsic differences in both dataset characteristics and methodological design.  

\paragraph*{Dataset Characteristics}  
In our experimental setting, the models are fine-tuned on the EarthReason dataset for reasoning segmentation and then transferred to three referring segmentation benchmarks: RRSIS-D, RISBench, and RefSegRS.  
These four datasets, in our opinion, collectively span a continuum from explicit referring to implicit reasoning, differing in both semantic granularity and geometric complexity.

RRSIS-D is a representative instance-level benchmark where each query uniquely identifies a single target instance, which often presents vast variations in scale and orientation. Its queries are therefore explicit, describing geometric and visual attributes (e.g., ``the gray rounded storage tank in the middle''). RISBench extends this instance-level paradigm with larger scale and linguistic diversity. It introduces longer and attribute-rich expressions with relational cues (e.g., ``the black vehicle at the middle-right edge of the image is partially hidden behind the house''). 

RefSegRS represents the early stage of referring segmentation task in remote sensing with a semantic-level paradigm. The task here is to segment all instances matching a semantic description, rather than a single object. Its expressions, generated from templates, are concise and specify an object category with an optional spatial context (e.g., ``building along the road''). This one-to-many grounding paradigm aligns with tasks requiring holistic scene understanding rather than single-object localization.

EarthReason generalizes the task into implicit semantic-level reasoning. Each query involves contextual or causal inference beyond direct visual naming. The conceptual landscape defined by these benchmarks illustrates a gradual shift along both visual and linguistic dimensions. This continuum transitions from geometric localization and attribute grounding to semantic abstraction and contextual reasoning, and defines the semantic–geometric landscape for our cross-dataset analysis.

\paragraph*{Methodological Design}  
The performance gap across datasets is also driven by the fundamental differences in model supervision.  
Think2Seg-RS is trained under a mask-only reinforcement learning paradigm, where the LVLM is optimized with a result-oriented reward signal centered on the final mask IoU. This learning scheme equips the model with strong \emph{semantic-level} reasoning and generalization abilities without relying on geometric priors such as bounding boxes or points.  
However, the absence of instance-level geometric priors fosters an inductive bias toward holistic semantic matching over pinpoint instance grounding.
In contrast, VisionReasoner and RemoteReasoner both adopt explicit \emph{instance-level} pseudo supervision, a strategy reliant on the costly generation of box–point prompts as spatial priors for each sample. 
This supervision strategy embeds a strong bias for instance-aware alignment, which directly matches the single-object localization objective of the RRSIS-D and RISBench benchmarks.

\paragraph*{Interpretation}  
Consequently, Think2Seg-RS achieves the best performance on \emph{semantic-level} datasets (EarthReason and zero-shot on RefSegRS), where semantic consistency and contextual reasoning dominate. Its policy learning directly aligns language-driven reasoning with holistic regional segmentation, producing coherent masks that match the implicit or explicit semantic intent of the query.
However, when transferred to \emph{instance-level} referring expression segmentation benchmarks (zero-shot on RRSIS-D, RISBench), the model's inductive bias toward holistic semantic matching conflicts with the core task of pinpointing a single instance, resulting in lower giou and P@0.5 than models trained with instance-aware geometric priors.
Nevertheless, Think2Seg-RS maintains a highly competitive cIoU on these benchmarks, demonstrating that while it may struggle to isolate small individual targets, it effectively captures the correct large-scale semantic regions.

These observed results therefore reflect not a weakness but a difference in \emph{inductive bias}: the mask-only signal nurtures a bias for holistic semantic understanding, whereas the box–point supervision signal fosters a strong bias toward precise geometric localization.
This divergence highlights a fundamental trade-off: Think2Seg-RS prioritizes semantic abstraction and context reasoning, while VisionReasoner and RemoteReasoner specialize in geometric grounding and fine spatial alignment.

\subsection{Ablation Study}

We conducted three ablation studies to evaluate the individual contributions of key components in the Think2Seg-RS framework.  
Specifically, we examined the influence of the LVLM prompter, the model size of SAM2, and various prompt combination strategies.  
All experiments were carried out on the EarthReason dataset, using Qwen-2.5-VL-3B as the default prompter and SAM2-Small as the default segmenter unless otherwise specified.
Note that while our main benchmark evaluations are reported as mean $\pm$ standard deviation across multiple runs to ensure statistical rigor, all ablation experiments were conducted using a single fixed random seed to maintain computational efficiency.

\subsubsection{Ablation on LVLM Prompter}  
The effect of different LVLM prompters and model sizes on the performance of Think2Seg-RS was evaluated.  
As presented in Table~\ref{tab:ablation_mllm}, multiple LVLM families and scales were compared, including LLaVA-1.5-7B~\citep{liu2023improvedllava}, the Qwen-2.5-VL series (3B and 7B)~\citep{bai2025qwen2}, and the InternVL3 series (2B and 8B)~\citep{zhu2025internvl3}.  
All experiments in this study employed SAM2-Small as the segmenter to ensure a consistent evaluation of the LVLM prompter’s influence.  
For the InternVL3 models, input images were resized to 896×896 pixels to match their default configuration.  

\begin{table}[htbp]
    \caption{Ablation study on LVLM prompter and model size}
    \label{tab:ablation_mllm}
    \centering
    \small
    \begin{tabular*}{0.48\textwidth}{@{\extracolsep{\fill}}lcccc}
    % \begin{tabular}{lcccc}
    \toprule
    \multirow{2}{*}{LVLM}   & \multicolumn{2}{c}{Validation}         & \multicolumn{2}{c}{Test}        \\
    \cmidrule(lr){2-3} \cmidrule(lr){4-5}
                            & gIoU           & cIoU           & gIoU           & cIoU           \\
    \midrule
    LLaVA-1.5-7B   & 38.60          & 35.84          & 39.26          & 36.94          \\
    InternVL3-2B            & 64.59          & 64.53          & 66.03          & 66.96          \\
    InternVL3-8B            & 72.16          & 71.91          & 73.24          & 73.13          \\
    Qwen-2.5-VL-3B          & 69.30          & 71.28          & 71.09          & 74.13          \\
    Qwen-2.5-VL-7B & \textbf{72.46} & \textbf{73.62} & \textbf{73.73} & \textbf{75.83} \\
    \bottomrule
    % \end{tabular*}
    \end{tabular*}
\end{table}

The results indicate that Think2Seg-RS achieves consistently strong performance when paired with recent high-capacity LVLMs, particularly the Qwen-2.5-VL and InternVL3 series.  
Among them, Qwen-2.5-VL-7B attains the overall best results on both validation and test sets, while InternVL3-8B also demonstrates comparable performance.  
A clear scaling trend is observed, where larger models within the same family outperform their smaller counterparts, suggesting that reasoning segmentation performance directly benefits from the enhanced reasoning and perception capabilities of larger-scale LVLMs.

It is also notable that LLaVA-1.5-7B performs significantly worse than the other LVLMs.  
This performance gap can be attributed to the architectural and training differences between LLaVA and other multimodal large language models such as Qwen-2.5-VL and InternVL3.  
LLaVA-1.5 was primarily optimized for visual dialogue and image captioning on natural images, relying on a limited visual encoder and a smaller, instruction-tuned dataset.  
In contrast, Qwen-2.5-VL and InternVL3 were trained with high-resolution perception and stronger multimodal alignment, making them substantially more capable of handling the complex spatial reasoning and semantic grounding required in reasoning segmentation tasks.  
Consequently, the results highlight the importance of both large-scale reasoning pretraining and fine-grained multimodal alignment for achieving high performance in Think2Seg-RS.

% revised
\begin{table}[!htbp]
    \caption{Ablation study on the model scale and variant of SAM2}
    \label{tab:ablation_sam2}
    \centering
    \small 
    \begin{tabular*}{0.48\textwidth}{@{\extracolsep{\fill}}lccccc}
    \toprule
    \multirow{2}{*}{SAM2 Variant} & \multirow{2}{*}{Params} & \multicolumn{2}{c}{Validation} & \multicolumn{2}{c}{Test} \\
    \cmidrule(lr){3-4} \cmidrule(lr){5-6}
    & & gIoU & cIoU & gIoU & cIoU \\
    \midrule
    SAM2-Tiny & 38.9M & 69.36 & 71.24 & 70.62 & 73.94 \\
    SAM2-Small & 46.0M & 69.30 & 71.28 & \textbf{71.09} & \textbf{74.13} \\
    SAM2-Base-plus & 88.8M & \textbf{69.50} & 70.38 & 70.78 & 72.84 \\
    SAM2-Large & 224.4M & 68.04 & 69.45 & 70.43 & 72.39 \\
    HQ-SAM2-Large & 224.7M & 69.22 & \textbf{71.54} & 70.63 & 73.03 \\
    \bottomrule
    \end{tabular*}
\end{table}

\subsubsection{Ablation on the Model Scale and Specialization of SAM2}  
To examine the influence of the segmentation backbone on overall performance, we evaluated the impact of both model scale and specialization of SAM2. We compared five variants: the standard SAM2 series (Tiny, Small, Base-plus, and Large) and a specialized variant, HQ-SAM2~\citep{sam_hq}, which is built upon the Large backbone and designed for high-quality segmentation. All other components were kept fixed. Unlike the LVLM ablation study, where larger models consistently improved reasoning capability, our results reveal a counterintuitive trend: increasing the capacity of the segmenter does not necessarily lead to better performance (Table~\ref{tab:ablation_sam2}). Specifically, while the specialized HQ-SAM2-Large improves upon the standard SAM2-Large—indicating that enhancing geometric priors is beneficial—it still falls short of the compact models. Among all variants, SAM2-Small achieves the highest test accuracy (71.09 gIoU and 74.13 cIoU) and the best trade-off between segmentation performance and computational efficiency, and is therefore adopted as the default segmenter in Think2Seg-RS.

\begin{figure}[!htbp]
    \centering
    \includegraphics[width=0.48\textwidth]{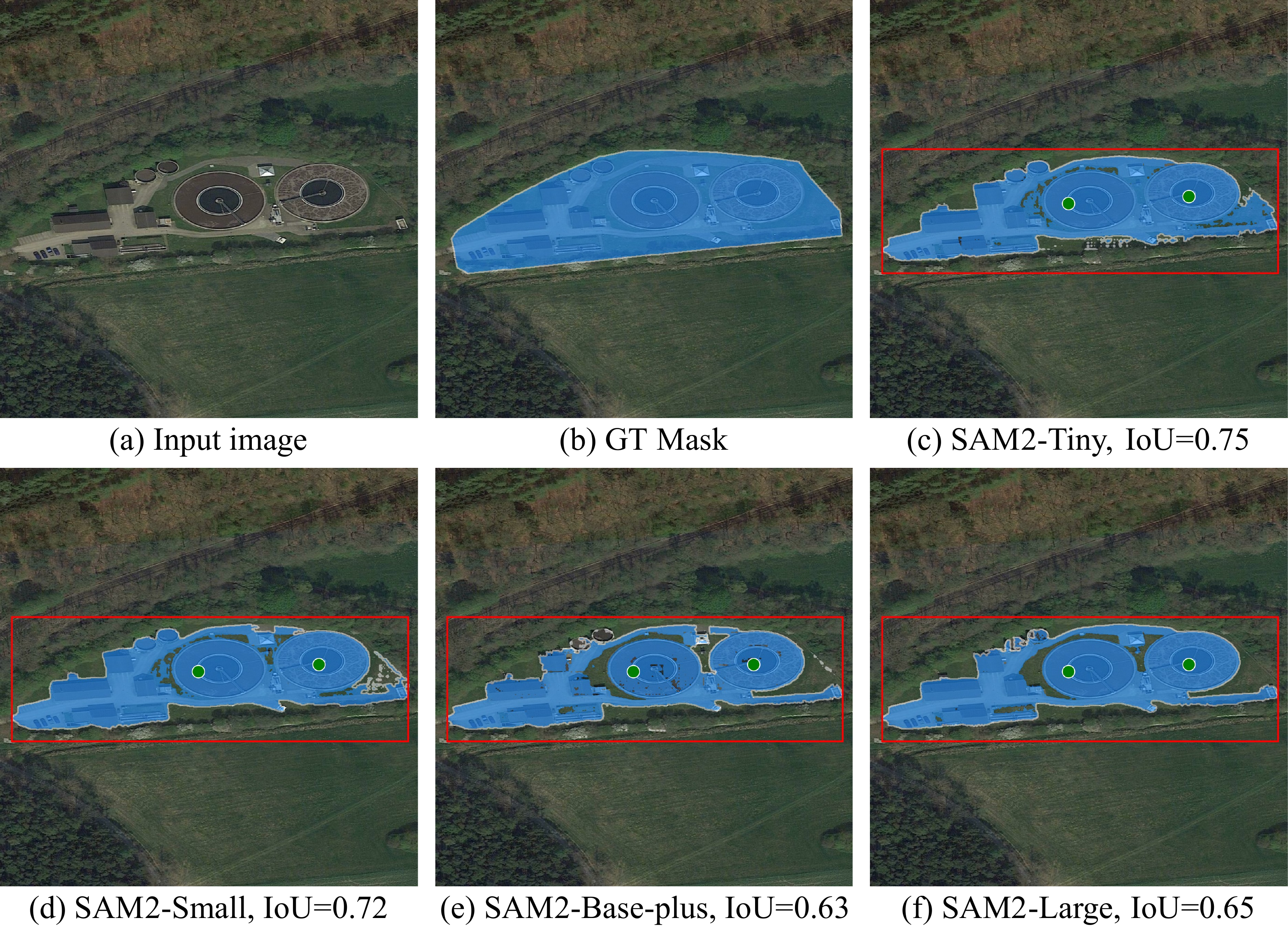}
    \caption{Effect of SAM2 model scale on semantic-level segmentation.  
    (a) Input image and (b) corresponding ground-truth (GT) mask illustrate a semantic-level annotation where the entire wastewater treatment plant is represented as one coherent polygon.  
    Outputs from four SAM2 variants, namely (c) Tiny, (d) Small, (e) Base-plus, and (f) Large, show that larger models generate over-detailed, fragmented masks misaligned with the coarse annotation style, resulting in lower IoU scores.  
    In contrast, smaller variants (Tiny and Small) produce smoother, spatially coherent masks that better match the semantic-level ground truth, achieving higher IoU values.  
    }
    \label{fig:sam_size_comparison}
\end{figure}

This phenomenon arises from the distinct interaction between segmentation granularity and the \emph{semantic-level annotation paradigm}.  
Larger variants like SAM2-Base-plus and SAM2-Large, including the specialized HQ-SAM2, possess fine-grained sensitivity that is advantageous for instance-level or natural-image tasks but becomes a liability in RS semantic-level grounding.
Although the HQ-SAM2 adaptation improves boundary precision over the standard Large model, it does not fundamentally resolve the semantic--geometric granularity mismatch.
These high-capacity models tend to over-segment by capturing small structural details and textural noise (e.g., variations in terrain, vegetation, or man-made substructures).  
As illustrated in Fig.~\ref{fig:sam_size_comparison}, when applied to a wastewater treatment plant, the larger models precisely delineate distinct man-made structures while strictly excluding the interstitial vegetation and ground. While these segmentations accurately identify meaningful sub-structures (i.e., individual objects within the facility), they result in a fragmented, quasi-instance-level output that poorly aligns with the unified, semantic-level ground truth mask.
Since the annotation is intentionally designed to represent the entire facility as a single coherent polygon, such excessive detail introduces a semantic mismatch that lowers the IoU score.  

Conversely, smaller variants like SAM2-Tiny and SAM2-Small exhibit an implicit regularization effect due to their constrained capacity.  
They smooth over fine textures and small gaps, producing spatially coherent masks that better match the coarse, semantic-level labeling prevalent in remote sensing.  
This alignment between model behavior and annotation granularity explains why smaller SAM2 variants outperform larger ones in reasoning segmentation.  
In essence, the optimal model scale is not absolute but depends on the desired level of semantic abstraction and the annotation style of the target domain.  
These findings empirically validate our design choice: for semantic-level reasoning tasks, a standard frozen segmenter (specifically the compact variant) offers a superior efficiency-accuracy trade-off compared to specialized high-precision variants, as the bottleneck lies in semantic grounding rather than geometric refinement.

% revised
\begin{table}[!htbp]
    \caption{Ablation Study on SAM Prompt Combination}
    \label{tab:ablation_prompt_combination}
    \centering
    \small
    \resizebox{0.99\columnwidth}{!}{
    \begin{tabular}{lcccc}
        \toprule
        \multirow{2}{*}{Prompt Combination}   & \multicolumn{2}{c}{val} & \multicolumn{2}{c}{test} \\
        \cmidrule(lr){2-3} \cmidrule(lr){4-5}
                                         & gIoU       & cIoU       & gIoU        & cIoU       \\
        \midrule
        Bbox Only                             & 67.61      & 69.68      & 68.27       & 70.95      \\
        Pos\_points$\times$2                    & 56.15      & 53.63      & 56.01       & 53.78      \\
        Bbox + Pos\_points$\times$2               & \textbf{69.30} & \textbf{71.28} & \textbf{71.09} & \textbf{74.13} \\
        Bbox + Pos\_points$\times$4               & 67.31      & 69.54      & 69.18       & 72.45      \\
        Bbox + Pos\_points$\times$2 + Neg\_points$\times$2 & 66.36      & 68.91      & 68.21       & 72.16      \\
        Autonomous Prompting         & 67.74      & 69.98      & 69.18       & 70.90       \\
        \bottomrule
    \end{tabular}
    }
\end{table}

\subsubsection{Ablation on Prompt Combination}
\label{sec:ablation_on_prompt_combination}

SAM is inherently sensitive to the type and configuration of input prompts.  
Since our Think2Seg-RS framework freezes the SAM segmenter and relies solely on the LVLM to generate prompts, the prompt combination becomes a critical factor that determines the final segmentation quality.  
To quantitatively and qualitatively understand this sensitivity, we conducted an extensive ablation study evaluating six configurations: (1) an autonomous prompting strategy where the LVLM is unconstrained to determine both the prompt combination and spatial placement, (2) bounding box only, (3) two positive points only, (4) bounding box with two positive points, (5) bounding box with four positive points, and (6) bounding box with two positive points plus two negative points.

\begin{figure*}[!htbp]
\centering
\includegraphics[width=0.90\textwidth]{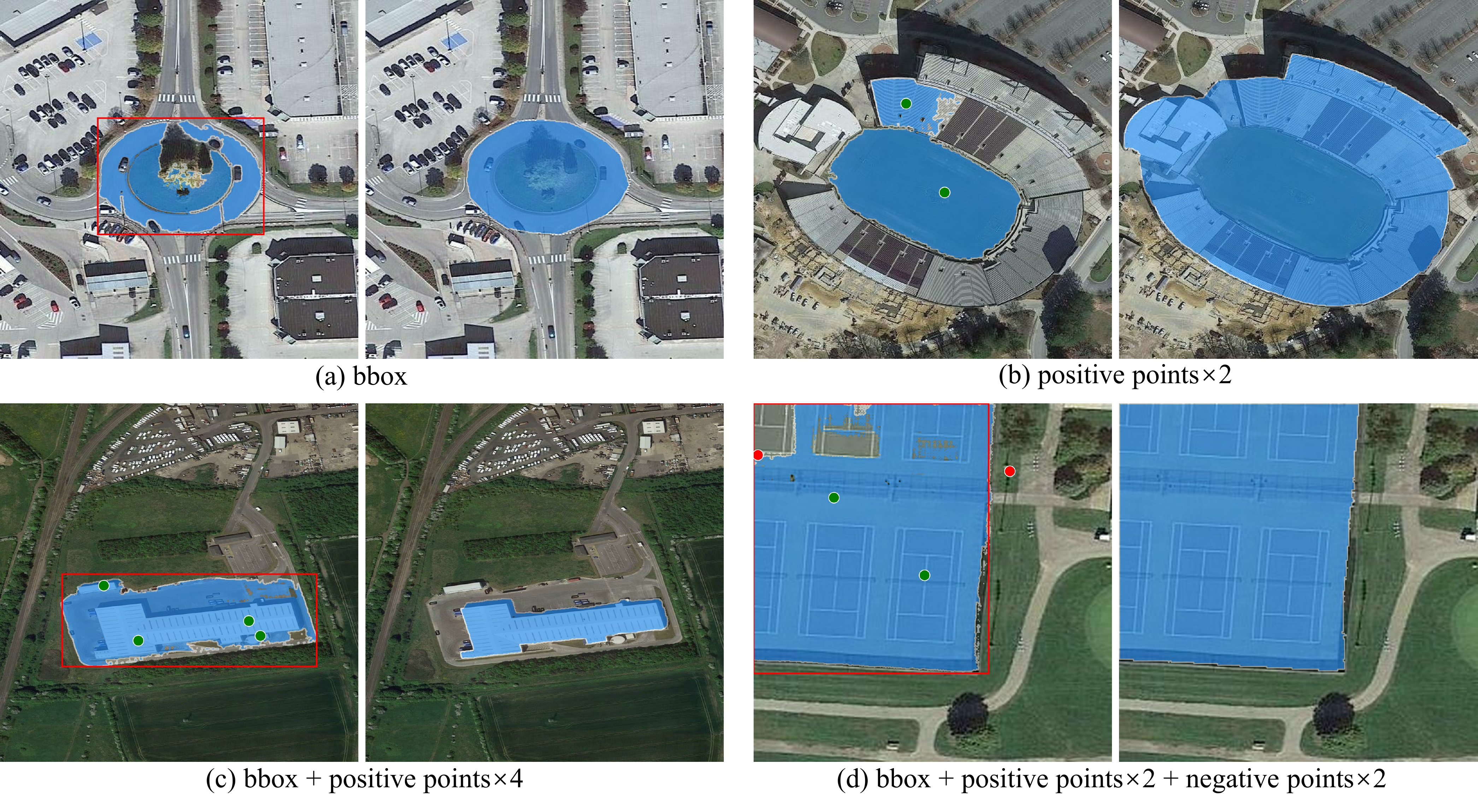}
    \caption{Qualitative analysis of prompting combination strategies for SAM. Each pair displays the input image with LVLM generated prompts and the resulting segmentation mask (left, bounding boxes in red, positive points in green, negative points in red) and the ground-truth mask (right). 
    }
    \label{fig:ablation_prompt_examples}
\end{figure*}

The quantitative results summarized in Table~\ref{tab:ablation_prompt_combination}, together with qualitative examples in Fig.~\ref{fig:ablation_prompt_examples}, provide a comprehensive view of the model’s behavior under different prompt combinations.
A single bounding box offers coarse localization but often covers multiple semantic regions. Without explicit semantic cues, the frozen SAM focuses on the dominant texture (e.g., pavement) and omits heterogeneous content (e.g., vegetation or shadows), producing an incomplete segmentation (Fig.~\ref{fig:ablation_prompt_examples}a).
Using only two positive points produces incomplete masks, as the limited cues cannot represent the diverse visual and semantic patterns within large or complex targets (Fig.~\ref{fig:ablation_prompt_examples}b).
Combining one bounding box with two positive points achieves the best overall performance by balancing semantic coverage and geometric constraint. The bounding box provides a global spatial prior, while the two positive points help disambiguate heterogeneous regions within the box, guiding SAM toward the intended semantic scope. This configuration serves as the default setting in Think2Seg-RS and consistently produces coherent regional masks across diverse scenes (see Fig.~\ref{fig:EarthReason_results}).

However, increasing the number of positive points to four does not further improve the results (Fig.~\ref{fig:ablation_prompt_examples}c).  
In many cases, these additional points fall into semantically diverse subregions within the same bounding box, such as rooftops, shadows, or adjacent ground textures, which causes the frozen SAM to merge inconsistent cues. This leads to overextended or irregular masks that blur the intended semantic scope.
Introducing negative points further degrades performance (Fig.~\ref{fig:ablation_prompt_examples}d).  
Unlike positive cues, which highlight coherent semantic areas, negative regions in aerial imagery are inherently heterogeneous, encompassing materials and textures such as vegetation, concrete, water, or asphalt.  
Such diversity prevents the LVLM from learning a stable placement strategy for negative prompts, and their inconsistent signals often conflict with positive cues. As a result, the generated masks become fragmented or distorted, reflecting the difficulty of defining uniform ``non-target'' semantics in high-resolution remote sensing scenes.

Finally, the autonomous prompting strategy provides critical validation for our fixed template design. Despite the theoretical flexibility to optimize prompt configurations, this unconstrained approach yields results virtually indistinguishable from the ``Bbox Only'' baseline, as the model empirically converges to generating exclusively bounding boxes. We attribute this observed mode collapse to the exponential complexity of the search space: the simultaneous optimization of prompt combinations (types/counts) and spatial geometries (positions/sizes) substantially exacerbates the difficulty of both maintaining format validity and maximizing the IoU reward. Consequently, driven by pre-training priors that favor bounding box grounding, Qwen-2.5-VL abandons the exploration of complex yet necessary fine-grained point interactions, confirming that a structured prompt template functions as a vital inductive bias to ensure effective policy learning.

Collectively, these findings underscore a fundamental characteristic of the decoupled LVLM--SAM architecture. 
Since the segmenter is frozen, all semantic information must be conveyed through spatial prompts generated by the LVLM. 
Consequently, prompt configuration directly determines how effectively high-level textual semantics are translated into low-level spatial cues. 
As evidenced by our ablation, when prompts are semantically well-aligned and structurally constrained (e.g., one box with two points), the interaction becomes efficient, enabling SAM to produce coherent regional masks. Conversely, misaligned, redundant, or unconstrained prompts distort this mapping, leading to under-segmentation, overextension, or mode collapse. 
These results highlight that the effectiveness of decoupled prompting depends not only on the reasoning ability of the LVLM but also on how efficiently the prompt space encodes semantic intent. 
This interplay between language-driven semantics and frozen visual priors defines the core behavior of the Think2Seg-RS framework.

\begin{table*}[htbp]
    \caption{Ablation study on training paradigms (SFT vs. GRPO) on the EarthReason dataset and zero-shot transfer benchmarks.}
    \label{tab:ablation_sft_vs_grpo}
    \centering
    \small
    \begin{tabular*}{\textwidth}{@{\extracolsep{\fill}}lcccccccccccccc}
        \toprule
        \multirow{3}{*}{Training Paradigm} & \multicolumn{4}{c}{EarthReason} & \multicolumn{9}{c}{Zero-shot Transfer Benchmarks} \\
        \cmidrule(lr){2-5} \cmidrule(lr){6-14}
         & \multicolumn{2}{c}{Validation} & \multicolumn{2}{c}{Test} & \multicolumn{3}{c}{RRSIS-D} & \multicolumn{3}{c}{RISBench} & \multicolumn{3}{c}{RefSegRS} \\
        \cmidrule(lr){2-3} \cmidrule(lr){4-5} \cmidrule(lr){6-8} \cmidrule(lr){9-11} \cmidrule(lr){12-14}
         & gIoU & cIoU & gIoU & cIoU & gIoU & cIoU & P@0.5 & gIoU & cIoU & P@0.5 & gIoU & cIoU & P@0.5 \\
        \midrule
        \multicolumn{14}{@{}l@{}}{\textit{Backbone: Qwen-2.5-VL-3B}} \\
        SFT & 65.83 & 67.08 & 67.30 & 67.97 & 26.78 & 30.27 & 26.20 & 36.24 & 32.40 & 35.40 & 1.52 & 3.35 & 0.28 \\
        GRPO & \textbf{69.30} & \textbf{71.28} & \textbf{71.09} & \textbf{74.13} & \textbf{43.42} & \textbf{48.66} & \textbf{45.10} & \textbf{40.04} & \textbf{36.49} & \textbf{39.37} & \textbf{14.91} & \textbf{16.46} & \textbf{6.60} \\
        \midrule
        \multicolumn{14}{@{}l@{}}{\textit{Backbone: Qwen-2.5-VL-7B}} \\
        SFT & 68.04 & 68.02 & 70.07 & 71.15 & 29.06 & 35.40 & 29.73 & 43.86 & 41.80 & 44.76 & 1.82 & 4.73 & 0.44 \\
        GRPO & \textbf{72.46} & \textbf{73.62} & \textbf{73.73} & \textbf{75.83} & \textbf{47.97} & \textbf{54.30} & \textbf{51.66} & \textbf{47.72} & \textbf{44.18} & \textbf{50.00} & \textbf{15.53} & \textbf{17.75} & \textbf{7.43} \\
        \bottomrule
    \end{tabular*}
\end{table*}

\subsubsection{Ablation on Training Paradigm: GRPO vs. SFT}
\label{sec:ablation_sft}

To explicitly verify the contribution of the reinforcement learning formulation, we compared our mask-only GRPO optimization against a Supervised Fine-Tuning (SFT) baseline. For this baseline, we employed the identical decoupled architecture, utilizing Qwen-2.5-VL as the prompter and a frozen SAM2-Small as the executor, but replaced the GRPO objective with a standard next-token prediction loss. Specifically, we utilized DBSCAN clustering~\citep{ester1996density} to decompose semantic masks into pseudo-instance targets. For each instance, we derived a fixed geometric prompt set comprising one bounding box and two positive points, strictly aligning with the output configuration of our RL-based model. Crucially, as ground-truth reasoning traces are unavailable, we excluded the intermediate Chain-of-Thought supervision, training the LVLM solely to map image–query pairs directly to the generated geometric coordinates, following the SFT baseline setup in \cite{shen2025vlm}.

The quantitative results are detailed in Table~\ref{tab:ablation_sft_vs_grpo}.
The GRPO paradigm consistently outperforms the SFT baseline across all metrics and model scales.
On the source dataset EarthReason, the GRPO method demonstrates superior performance, with the 7B model achieving gains of +3.66 gIoU and +4.68 cIoU on the test set compared to SFT.
Crucially, this performance gap is further magnified on the out-of-domain zero-shot benchmarks; notably, on the semantic-level RefSegRS dataset, the SFT paradigm collapses almost completely, whereas GRPO maintains robust zero-shot performance.
We attribute this disparity to the fundamental difference in optimization objectives between the two paradigms: the SFT approach, restricted to minimizing the next-token prediction loss, tends to rely on the memorization of specific answers (i.e., the fixed pseudo-labels). In contrast, driven by the goal of maximizing the final reward, the GRPO method is incentivized to explore and identify superior prompting strategies, thereby fostering the generalization of the geometric reasoning policy. 
These results corroborate recent studies suggesting that RL is essential for learning transferable principles beyond rote memorization~\citep{chu2025sft}.

\subsection{Computational Cost Analysis}
\label{sec:compute_efficiency}

In addition to accuracy, we evaluate the computational cost of Think2Seg-RS in Table~\ref{tab:efficiency}. To ensure a fair comparison, all models were tested on identical NVIDIA A800 (80GB) GPUs, with reported statistics averaged over multiple independent runs to ensure reliability.

\begin{table}[htbp]
    \centering
    \small
    \begin{threeparttable}
    \caption{Computational cost analysis on NVIDIA A800 (80GB).}
    \label{tab:efficiency}
    \begin{tabular*}{0.48\textwidth}{@{\extracolsep{\fill}}lccc}
        \toprule
        Method & Time (h) & \#GPUs & Inf. Mem. (GB) \\ 
        \midrule
        SegEarth-R1$^*$ & N/A & N/A & 6.96 \\
        RemoteReasoner$^*$ & N/A & N/A & 19.50 \\
        Seg-Zero$^\dag$ & 54.0 & 4 & 17.34 \\
        VisionReasoner$^\dag$ & 52.7 & 4 & 16.90 \\ 
        Think2Seg-RS (3B) & 42.7 & 2 & 7.86 \\
        Think2Seg-RS (7B) & 53.0 & 4 & 16.30 \\ 
        \bottomrule
    \end{tabular*}
    \begin{tablenotes}
        \footnotesize
        \item \textit{Note:} $^*$ Based on official paper/checkpoints. $^\dag$ Reproduced using our experimental setting.
    \end{tablenotes}
    \end{threeparttable}
\end{table}

As shown in Table~\ref{tab:efficiency}, the training time of Think2Seg-RS (7B) is comparable to other GRPO-based decoupled frameworks , requiring approximately 53.0 hours on four GPUs. During inference, owing to the adoption of the compact SAM2-Small segmenter, Think2Seg-RS (7B) requires 16.30 GB of memory, maintaining a slight efficiency advantage over its counterparts. Notably, the 3B variant consumes only 7.86 GB, offering a lightweight deployment option comparable to the end-to-end SegEarth-R1 (6.96 GB).

\section{Discussion}

\paragraph{Scope and positioning}
Think2Seg-RS reframes reasoning segmentation in remote sensing as a process of learning to prompt a frozen segmenter.  
By optimizing an LVLM through a lightweight, mask-only GRPO signal, the framework transfers high-level reasoning into geometric prompts that SAM2 can reliably execute.  
Here we discuss the broader implications of our findings, focusing on cross-task applicability, sensitivity of the prompt interface, and the interaction between segmenter capacity and annotation granularity.

\paragraph{Cross-task generalization and applicability domain} 
Our zero-shot transfer experiments revealed a consistent pattern: Think2Seg-RS generalizes strongly to \emph{semantic-level referring tasks}, yet shows a performance gap on \emph{instance-level} benchmarks. This divergence stems fundamentally from the topological invariance of the global IoU metric employed in our reward function.
While optimizing for global IoU ensures semantic coherence and scene-level completeness, which are advantageous for EarthReason and RefSegRS, this formulation highlights an inherent limitation: it prioritizes holistic semantic coverage over discriminative instance isolation, as the metric aggregates pixel overlap irrespective of the number of connected components. Consequently, the objective function lacks the explicit incentive required for instance separation, leading to under-segmentation in datasets like RRSIS-D and RISBench where adjacent targets must be individually grounded. 
This behavior is statistically evidenced across multiple independent runs (Table~\ref{tab:earthreason_results} and ~\ref{tab:rrsis_zero-shot_results}) by the disproportionate gains in cIoU relative to gIoU: the former measures overall semantic area coverage via dataset-level pixel aggregation, whereas the latter evaluates average per-sample geometric accuracy through equal sample weighting.
Bridging this gap calls for incorporating separation-aware rewards without reverting to costly box supervision. A promising strategy is to introduce a connectivity consistency term that penalizes the discrepancy in the number of connected components between the predicted mask and the ground truth, thereby enforcing instance-level count fidelity. Furthermore, implicit geometric attributes of the ground-truth mask, such as centroids or principal axes, could serve as weak supervisory signals to guide the LVLM in generating spatially distinct prompts for clustered objects. These augmentations would effectively enhance instance precision while preserving the strong semantic-level generalization that characterizes Think2Seg-RS.

\paragraph{Prompt interface sensitivity and the role of negative points}
Because the segmenter is frozen, all semantic information is conveyed through the spatial prompts generated by the LVLM.  
Our ablation results show that this interface performs best when balancing semantic coverage and geometric constraint (one box with two positive points), but becomes unstable when prompts are sparse, redundant, or semantically conflicting.  
In particular, while negative prompting is a potent mechanism in interactive segmentation for background suppression, its underperformance here highlights a challenge in the automated RL formulation.
The intrinsic heterogeneity of non-target regions in aerial imagery (e.g., vegetation, roofs, asphalt) creates a noisy optimization landscape.
Without a supervised warm-up (SFT cold start) or explicit geometric constraints (e.g., a penalty term forbidding negative points falling within the target region), the LVLM struggles to learn a consistent policy for placing negative points that effectively suppress distractors.
This suggests that the limited impact observed in our study is not a failure of negative prompting, but a difficulty in learning optimal placement strategies. 
This observation aligns with prior remote sensing studies leveraging SAM: the SAMRS dataset~\citep{wang2023samrs} excluded background points, noting the difficulty of accurately defining them in an automated pipeline, while ~\cite{chen2024rsprompter,zhang2023text2seg,osco2023segment} also rely on boxes or positive points, bypassing the use of negative points for object localization.
Future work could address this by incorporating a brief supervised fine-tuning stage or designing auxiliary rewards that explicitly guide the policy to target look-alike distractors.

\paragraph{From fixed schema to adaptive prompting policies}
While an adaptive policy that autonomously determines the number and type of prompts is theoretically ideal, our empirical findings in the ablation study~\ref{sec:ablation_on_prompt_combination} demonstrate the necessity of a fixed schema to stabilize training.
Unconstrained optimization suffers from mode collapse, where the LVLM reverts to a ``safe'' local optimum of simple box generation due to the intractable search space and pre-training priors. The fixed schema thus acts as a necessary inductive bias, pruning the action space and compelling the model to master fine-grained point placement. 
However, this structural constraint inherently limits the model's flexibility when handling targets with complex or irregular geometries.
A key future objective is to realize a fully adaptive policy that selects optimal templates conditioned on inferred target morphology and background heterogeneity (see Fig.~\ref{fig:prompt_combination_and_target_characteristics} for illustrative examples).
To mitigate mode collapse in this unconstrained setting, future research should explore complexity-aware reward shaping. By modulating the reward signal with target characteristics (e.g., shape irregularity or background clutter), the policy can be incentivized to deploy sophisticated prompt combinations solely when necessitated by geometric complexity, thereby reconciling flexibility with optimization stability.

\paragraph{Segmenter scale versus annotation granularity}
A counterintuitive but consistent observation is that smaller SAM2 variants outperform larger ones under semantic-level supervision.  
High-capacity segmenters tend to overfit to high-frequency textures and microstructures, which misalign with the coarse semantic masks provided in reasoning segmentation.  
In contrast, compact models act as implicit regularizers, smoothing over noisy details and better capturing the primary object structures relevant to semantic-level tasks. 
This observation suggests a general principle for the RS domain: the capacity and inductive bias of the execution engine should match the granularity and abstraction level of the annotation.
Such alignment yields better accuracy–efficiency trade-offs and avoids unnecessary scaling in large remote-sensing models.

\paragraph{Limitations and actionable next steps}
Think2Seg-RS currently relies on a single-turn policy and a fixed prompt schema, without explicit uncertainty estimation.  
Several extensions are immediately actionable:  
(i) introducing geometry-aware rewards that incorporate orientation, skeleton, and length consistency for slender or rotated targets;  
(ii) adopting adaptive prompting strategies that condition the prompt template on inferred target morphology and background heterogeneity, including semantic-aware negatives; and  
(iii) integrating minimal calibration data that mixes limited referring-style supervision or multi-turn corrections to refine instance precision while maintaining scalability.  
These steps preserve the advantages of decoupling, including modularity, efficiency, and interpretability, while broadening the applicability from semantic-level reasoning toward fine-grained, instance-centric grounding.

\paragraph{Takeaway}
The observed transfer patterns and ablation behaviors do not weaken the contribution of Think2Seg-RS; rather, they clarify its applicability domain and illuminate clear directions for extension.  
Mask-only GRPO equips the LVLM with a transferable prompting skill that excels in tasks emphasizing semantic coherence.  
By enriching the reward and prompt space with lightweight geometric priors and adaptive policies, the same decoupled formulation can be extended toward high-precision referring segmentation in instance-dense remote-sensing scenes.

\section{Conclusion}

In this study, we presented Think2Seg-RS, a decoupled large vision–language framework for reasoning segmentation in remote sensing.  
By training an LVLM prompter to guide a frozen SAM2 segmenter through geometric prompts and optimizing it with a mask-only GRPO reward, Think2Seg-RS bridges abstract semantic reasoning with concrete spatial delineation.  
This paradigm achieves state-of-the-art performance on the EarthReason dataset and exhibits strong zero-shot transfer to referring segmentation benchmarks, demonstrating robust cross-dataset and cross-task generalization.

Experiments further reveal that compact SAM2 variants better align with semantic-level supervision, while negative prompting remains challenging due to the heterogeneous backgrounds in aerial imagery. 
Overall, Think2Seg-RS provides an efficient and scalable paradigm for semantic-level reasoning segmentation and points toward future extensions in adaptive prompting, geometry-aware rewards, and unified semantic–instance segmentation for geospatial analysis.

\section*{Declaration of Competing Interest}

The authors declare that they have no known competing financial interests or personal relationships that could have appeared to influence the work reported in this paper.

\section*{Declaration of generative AI and AI-assisted technologies in the manuscript preparation process}

During the preparation of this manuscript, the authors utilized ChatGPT to assist with language enhancement and improve readability. After utilizing this tool, the authors thoroughly reviewed and revised the content as necessary, and take full responsibility for the final manuscript.

\section*{Acknowledgments}
This work was supported in part by the National Natural Science Foundation of China (Grant Nos. 62476205 and 62301405), and the Fundamental Research Funds for the Central Universities (Grant Nos. ZYTS25152 and ZYTS25136).

% \nolinenumbers

% \small
\bibliographystyle{elsarticle-harv} 
\bibliography{reference_revised}

\clearpage

\appendix
\renewcommand{\thefigure}{\thesection.\arabic{figure}}
\makeatletter
\@addtoreset{figure}{section}
\makeatother

\section{Instruction Template for the LVLM}
\label{sec:A1}

\begin{figure}[ht]
    \centering
    \includegraphics[width=0.45\textwidth]{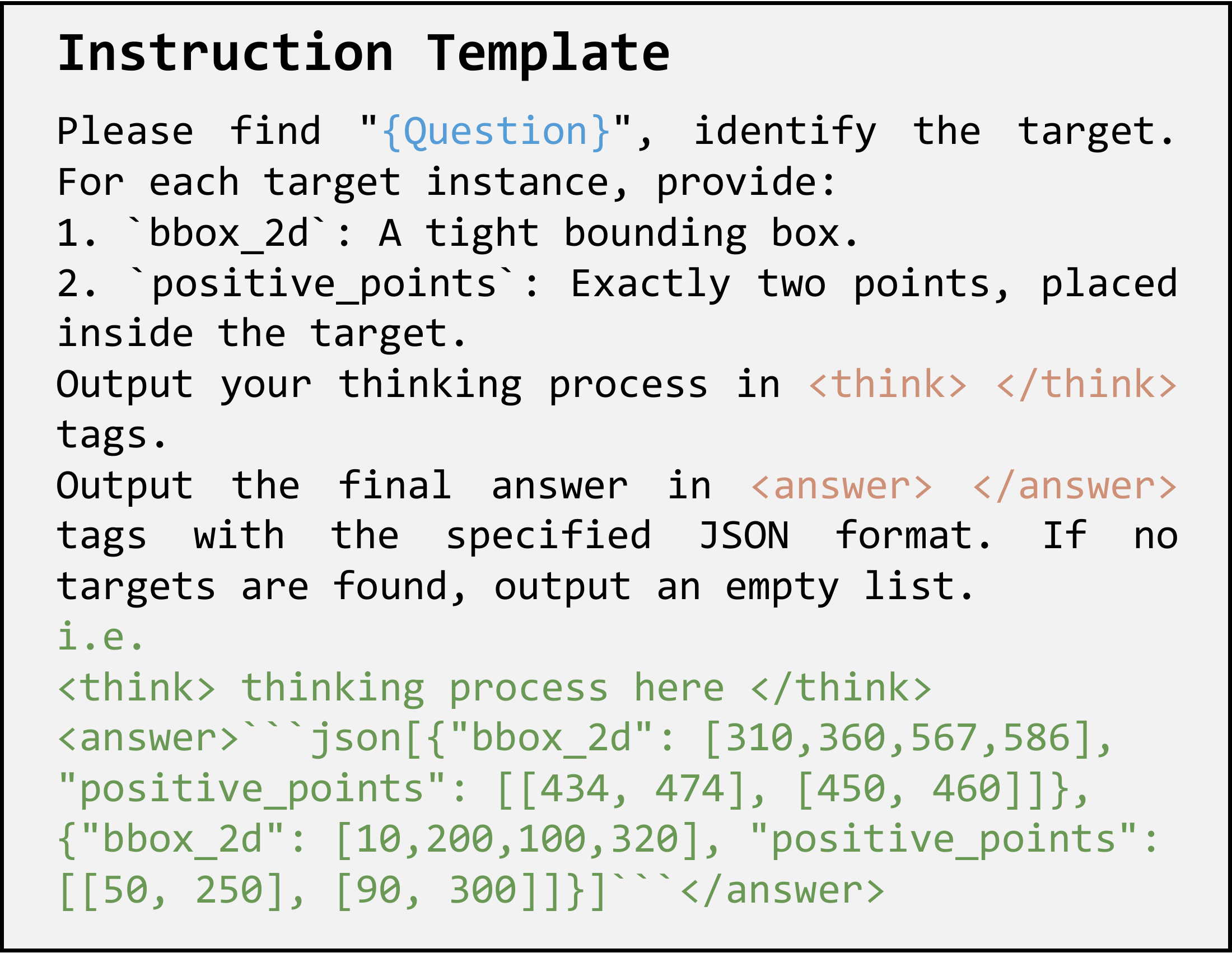}
    \caption{Structured instruction template for the LVLM. The template defines the task, the CoT reasoning format, and the required JSON schema for geometric prompts. 
    The \texttt{\{Question\}} placeholder is dynamically replaced with the implicit, complex reasoning query for each input sample, ensuring that the LVLM generates consistent reasoning traces and well-formed prompts for downstream SAM execution.}
    \label{fig:prompt_template}
\end{figure}

To ensure that the LVLM produces consistent and correctly formatted outputs, we design a structured instruction template. As illustrated in Fig.~\ref{fig:prompt_template}, this template guides the LVLM by explicitly defining (i) the task to be performed, (ii) the requirement to generate a CoT reasoning process enclosed in \verb|<think>| tags, and (iii) the JSON schema for the visual (geometric) prompts enclosed in \verb|<answer>| tags. The generated geometric prompts are then passed to SAM for execution, making this template a critical component for the automated reasoning–execution pipeline.

\section{Supplementary Analysis and Visualizations}

This appendix presents supplementary analysis and visualizations to provide a broader and more nuanced view of the Think2Seg-RS framework. Complementing the quantitative findings in the main paper, these results offer deeper insights into the model's behavior, strengths, and potential for future adaptive strategies.

\subsection{Additional Qualitative Results}

\begin{figure*}[htbp]
\centering
\includegraphics[width=0.95\textwidth]{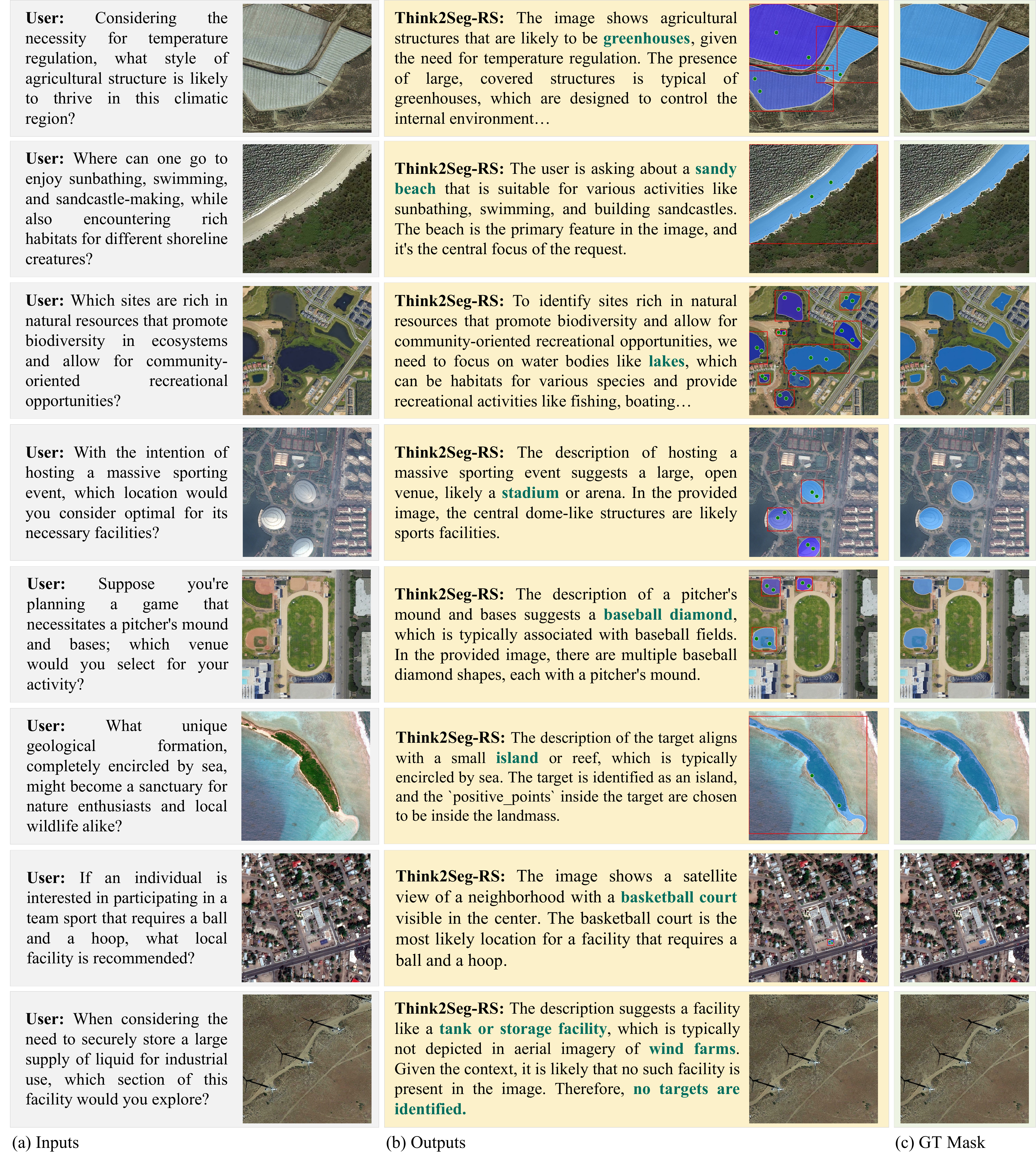}
\caption{Additional qualitative results of Think2Seg-RS on the EarthReason dataset. 
(a) Inputs, including the user query and corresponding remote sensing image. 
(b) Model outputs, comprising the LVLM's reasoning text from the \texttt{<think>} stage, the generated geometric prompts (bounding box in red and positive points in green), and the resulting segmentation mask predicted by SAM2. 
(c) Ground-truth (GT) mask for reference.}
\label{fig:EarthReason_results_appendix}
\end{figure*}

We first display additional results on the EarthReason dataset (Fig.~\ref{fig:EarthReason_results_appendix}), illustrating the model’s robustness in handling complex and implicit queries.  
These examples highlight how the framework effectively bridges high-level semantic reasoning with the generation of precise geometric prompts.

\subsection{Failure Case Analysis}
To rigorously understand the limitations of our framework, we conducted manual inspection of all 378 validation samples where the Think2Seg-RS-7B model failed to achieve a satisfactory segmentation (gIoU $< 0.5$). We categorized the error sources into four distinct types:
\begin{itemize}
    \item \textbf{Reasoning Failure:} The LVLM misinterprets the high-level semantic constraints or logic of the query, leading to the identification of an incorrect target object. This definition also extends to the misjudgment of target presence, encompassing scenarios where the model fails to reject invalid queries or overlooks existing targets.
    \item \textbf{Prompting Failure:} The LVLM correctly identifies the semantic target in its textual reasoning but fails to translate this understanding into accurate geometric prompts. This manifests as either spatial misalignment (e.g., grounding on a visually similar distractor or a shadow) or geometric imprecision (e.g., generating loosely fitted bounding boxes) that leads to degraded segmentation quality.
    \item \textbf{Execution Failure:} The LVLM provides accurate reasoning and well-placed prompts, but the frozen SAM segmenter generates suboptimal masks, often due to a ``granularity gap'' between instance-level perception and semantic-level annotation.
    \item \textbf{Annotation Inconsistency or Semantic Ambiguity:} This type represents cases where the model's prediction is logically valid but differs from the ground truth due to the inherent ambiguity of high-level reasoning tasks or subjective annotation standards. Unlike the first three categories, these are not intrinsic model errors but highlight the inherent challenge of establishing a unique ground truth for open-ended reasoning tasks.
\end{itemize}

\begin{figure*}[htbp]
\centering
\includegraphics[width=0.95\textwidth]{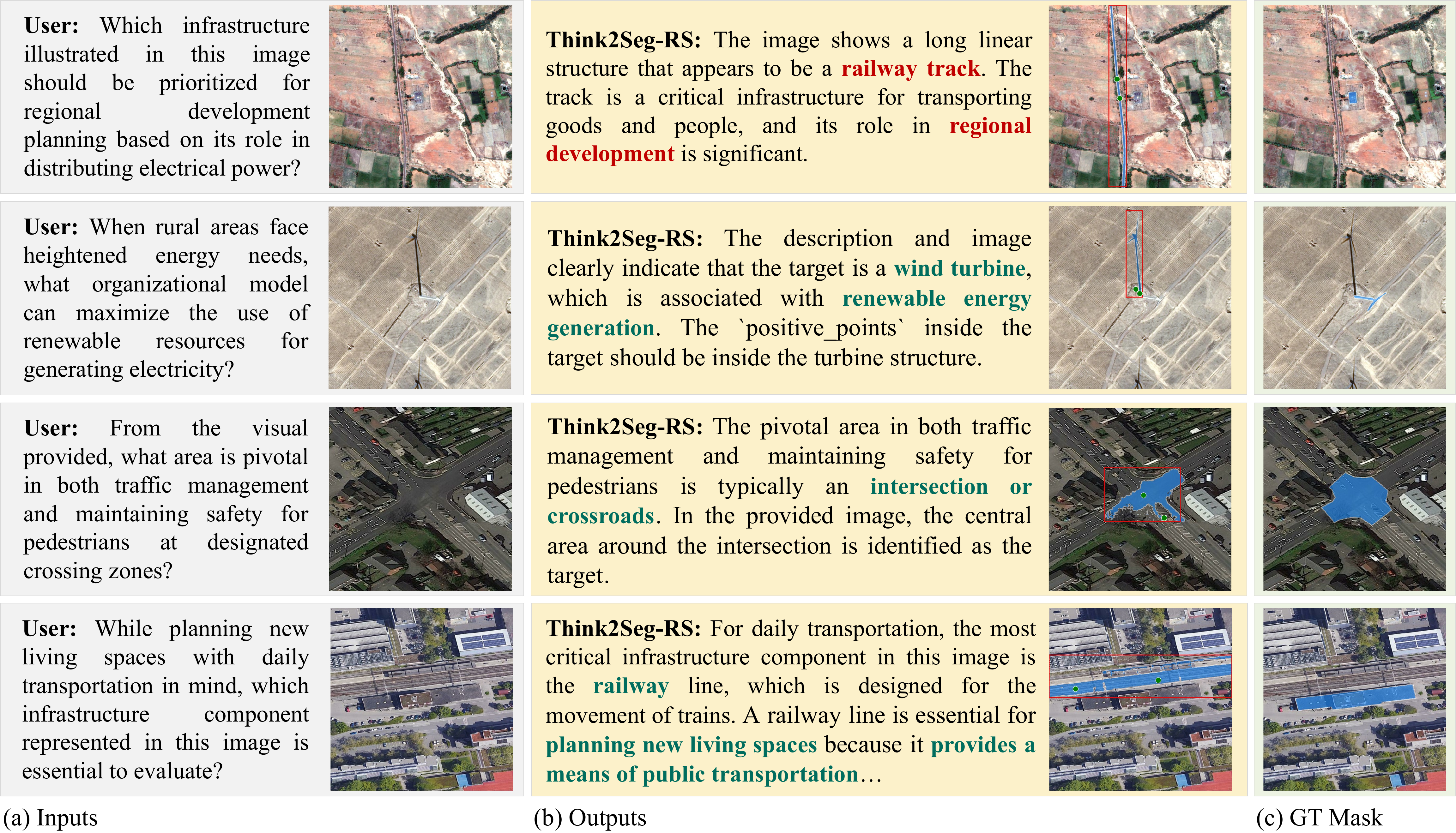}
\caption{Failure cases of Think2Seg-RS on the EarthReason dataset. 
(a) Inputs, including the user query and corresponding remote sensing image. 
(b) Model outputs, comprising the LVLM's reasoning text from the \texttt{<think>} stage, the generated geometric prompts (bounding box in red and positive points in green), and the resulting segmentation mask predicted by SAM2. 
(c) Ground-truth (GT) mask for reference.}
\label{fig:failure_case}
\end{figure*}

To provide a concrete understanding of these failure types, we analyze representative cases shown in Fig.~\ref{fig:failure_case}. 
The first row illustrates a high-level reasoning failure, where the LVLM misinterprets subtle query constraints, prioritizing the general concept of ``regional development'' over the specific function of ``distributing electrical power,'' leading it to misidentify a railway line as the target instead of the electrical substation. 
The second row demonstrates a prompting failure caused by visual distractors. Although the model correctly identifies the target as a ``wind turbine'', the prominent cast shadow misleads the LVLM to place the grounding prompt on the shadow rather than the turbine structure itself. 
The third row highlights an execution failure stemming from the intrinsic limitations of SAM. Even with accurate reasoning and well-positioned prompts for the ``intersection,'' the target area lacks distinct boundary definitions or specific textural patterns to separate it from the surrounding road network. Consequently, the frozen SAM—which relies on explicit visual features for object delineation—struggles to define the extent of the region, resulting in a fragmented mask that fails to cover the holistic functional area defined in the ground truth.
Finally, the fourth row exemplifies annotation inconsistency. The user queries for an infrastructure component for ``daily transportation,'' where both the ``railway line'' (identified by the model) and the ``train station'' (defined in the ground truth) are valid interpretations. The model's prediction is logically sound but penalized by the specific, subjective definition of the ground truth.

We present the statistical distribution of these failure modes in Table~\ref{tab:failure_breakdown}. The quantitative results highlight the current bottlenecks and priority areas for future improvement.
The statistical analysis identifies prompting failure as the dominant error source, constituting nearly 60\% of all failure cases. This finding pinpoints the translation from abstract semantic reasoning to precise spatial coordinates as the primary bottleneck, suggesting that future research should prioritize robust geometric grounding to generate high-precision cues. 
Execution failure, the second most frequent mode, underscores the inherent limitations of employing a frozen, general-purpose segmenter. Specifically, it exposes the ``granularity gap'' where SAM’s texture-dependent perception diverges from the broader semantic-level regions defined by annotations. 
Conversely, the remarkably low incidence of reasoning failure validates the strong interpretive capacity of the LVLM, confirming its proficiency in handling implicit geospatial logic. 
Finally, the minor fraction attributed to semantic ambiguity reflects the ontological challenges inherent in open-world reasoning, where valid alternative interpretations may differ from rigid ground-truth standards despite being logically sound.

\begin{table}[htbp]
    \centering
    \small
    \caption{Quantitative breakdown of failure modes on the subset of unsuccessful validation samples (N=378).}
    \label{tab:failure_breakdown}
    \begin{tabular*}{0.48\textwidth}{@{\extracolsep{\fill}}lcc}
    % \begin{tabular*}{0.48\textwidth}{lcc}
        \toprule
        \textbf{Failure Mode} & \textbf{Count} & \textbf{Percentage}  \\
        \midrule
        Prompting Failure & 225 & 59.5\%   \\
        Execution Failure & 114 & 30.2\%   \\
        Reasoning Failure & 26 & 6.9\%   \\
        Semantic Ambiguity & 13 & 3.4\% \\
        \bottomrule
    \end{tabular*}
\end{table}

\subsection{Analysis of Adaptive Prompting Strategies}

\begin{figure*}[htbp]
    \centering
    \includegraphics[width=0.90\textwidth]{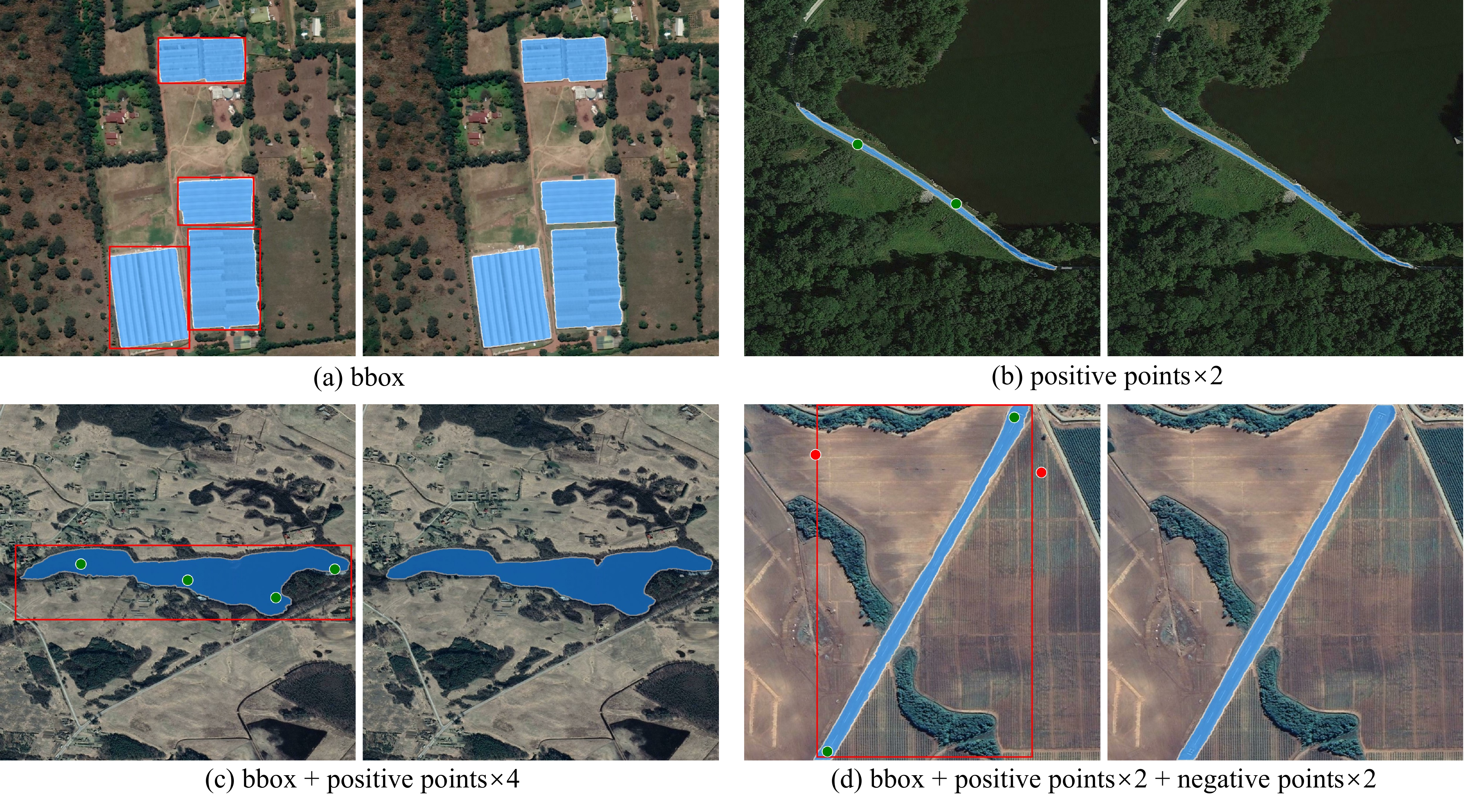}
    \caption{Examples of prompt combinations tailored to different target characteristics. 
    Each pair displays the input image with generated prompts and the resulting segmentation mask (left, bounding boxes in red, positive points in green, negative points in red) and the ground truth mask (right). 
    (a) For targets with regular, box-like geometries such as the greenhouses shown, a simple bounding box is effective and sufficient. 
    (b) When segmenting thin, linear features like the narrow dam, two positive points placed along the object's axis can define its trajectory precisely. 
    (c) For objects with irregular and complex boundaries like the lake, a bounding box combined with multiple positive points is crucial for capturing the intricate details of the boundary. 
    (d) In scenarios where a slender target occupies a small fraction of a bounding box, negative points become essential for explicitly excluding the large, visually similar background, thus preventing oversegmentation.}
    \label{fig:prompt_combination_and_target_characteristics}
\end{figure*}

Finally, Fig.~\ref{fig:prompt_combination_and_target_characteristics} illustrates how the optimal prompt configuration varies with the geometric characteristics of the target.  
Although this adaptive selection mechanism is not yet implemented in the current framework, the examples provide an intuitive visualization of the direction discussed in our future work, where the LVLM could first infer the target’s structural type (e.g., regular, elongated, or irregular) and then dynamically choose the most suitable combination of bounding boxes and points.  
This highlights the potential of extending Think2Seg-RS toward adaptive prompting strategies that generalize across diverse target geometries.

% %% For authoryear reference style

% \bibitem[Lamport(1994)]{lamport94}
%   \textit{\LaTeX: a document preparation system},
%   Addison Wesley, Massachusetts,

% \end{thebibliography}
\end{document}